\documentclass[a4paper,10pt]{article}
\usepackage{ifthen}
\newboolean{review}
\setboolean{review}{false}

\usepackage[nobiblatex]{xurl}
\usepackage[breaklinks,colorlinks]{hyperref}
\usepackage{doi} 
\usepackage[hyperref=true,natbib,backend=bibtex,style=ext-authoryear-comp,maxnames=2,giveninits=true,dashed=false]{biblatex}
\DeclareFieldFormat{labelnumberwidth}{#1\adddot}
\DeclareNameAlias{sortname}{family-given}
\DeclareNameAlias{author}{family-given} 
\DefineBibliographyStrings{english}{and = {\&}} 
\DeclareFieldFormat
  [misc,article,inbook,incollection,inproceedings,unpublished]
  {titlecase:title}{\MakeSentenceCase*{#1}} 
\DeclareFieldFormat[misc,article,inbook,incollection,inproceedings,thesis,book,unpublished]{title}{#1} 
\renewbibmacro{in:}{%
  \ifboolexpr{%
     test {\ifentrytype{article}}%
     or
     test {\ifentrytype{inproceedings}}%
  }{}{\printtext{\bibstring{in}\intitlepunct}}%
} 
\AtEveryBibitem{\clearname{editor}} 
\DeclareFieldFormat[article]{volume}{#1\addcolon\space}
\DeclareFieldFormat[article]{pages}{#1} 
\AtEveryBibitem{%
  \ifentrytype{article}{%
    \clearfield{number}%
    \clearfield{issue}%
  }{}%
} 

\usepackage[dvipsnames]{xcolor}
\hypersetup{colorlinks=true,citecolor=blue,urlcolor=blue,linkcolor=blue,menucolor=blue}

\usepackage[capitalize]{cleveref}

\usepackage{afterpage} 

\usepackage{graphicx} 
\graphicspath{{Figs/}}

\usepackage{subfigure} 
\usepackage[bf]{caption} 
\usepackage{multirow}
\usepackage{rotating}

\usepackage{siunitx,array,booktabs,ragged2e}
\newrobustcmd\B{\DeclareFontSeriesDefault[rm]{bf}{b}\bfseries}
\newrobustcmd\W{\DeclareFontSeriesDefault[rm]{bf}{b}\bfseries\color{Maroon}}
\newrobustcmd\F{\DeclareFontSeriesDefault[rm]{bf}{b}\bfseries\color{MidnightBlue}}

\usepackage[frozencache]{minted}
\usepackage{algorithm}

\usepackage[a4paper,portrait,margin=2cm]{geometry}

\usepackage{times} 
\usepackage{amsfonts}   
\usepackage{dsfont}
\usepackage{nicefrac}       
\usepackage{microtype}      

\usepackage[nodayofweek]{datetime} 

\usepackage{lineno}
\usepackage{enumitem} 
\setlist[itemize]{topsep=0pt,parsep=0pt,itemsep=0pt,leftmargin=1.8em}
\setlist[enumerate]{topsep=0pt,parsep=0pt,itemsep=0pt,leftmargin=2.4em}

\usepackage{multicol}

\usepackage{xspace}
\newcommand{\etc}{{\it etc.}\xspace}
\newcommand{\ie}{{\it i.e.},\xspace}
\newcommand{\eg}{{\it e.g.},\xspace}

\newcommand{\hide}[1]{}

\makeatletter
  \renewcommand\maketitle{\hfill \titlebibdata\\\par\noindent {\Large \bf \@title}\\[-1mm]\par \noindent {\large \bf \authorA}\hfill {\tt \emailA}\\[0.8mm] \noindent {\large \bf \authorB}\hfill {\tt \emailB}\\[0.8mm]\address}
\makeatother

\begin{document}


\newcommand{\titletext}{Redesigning Neural Networks to Not be Confidently Wrong even to Strong Adversarial Attacks}

\title{\titletext}
\newcommand{\authorA}{Michael W. Spratling}
\newcommand{\authorB}{Heiko H. Sch{\"u}tt}
\newcommand{\address}{University of Luxembourg, Department of Behavioural and Cognitive Sciences, L-4366 Esch-sur-Alzette, Luxembourg.}
\newcommand{\addressfull}{Department of Behavioural and Cognitive Sciences\\University of Luxembourg\\Maison Sciences Humaines\\11 Porte des Sciences\\L-4366 Esch-sur-Alzette\\Luxembourg}

\newcommand{\emailA}{{\tt michael.spratling@uni.lu}}
\newcommand{\emailB}{{\tt heiko.schutt@uni.lu}}

\newcommand{\orcidA}{0000-0001-9531-2813}
\newcommand{\orcidB}{0000-0002-2491-5710}
\newcommand{\titlebibdata}{\vspace*{-2ex}} 

\maketitle

\ifreview
\linenumbers
\fi

\begin{abstract}%
Early deep learning models attempted to simulate visual perception using a neural hierarchy that alternated between two types of layer: one modelled on cortical simple cells (that was designed to increase selectivity), and one modelled on cortical complex cells (that aimed to increase tolerance). We show that such a neural network (NN) architecture, when implemented with novel components that are better able to learn selective and tolerant representations, is capable of accurate and robust image classification and improved alignment between the behaviour of the model and the perceptual abilities of humans.
Firstly, the proposed models are shallow, fewer than ten layer-pairs, which is more consistent with the rapid feedforward object categorisation performed by the biological visual system.
Secondly, they are far less prone to producing classification predictions with high confidence for samples from unknown classes or for image consisting of pure noise, and hence, like humans are not easily fooled into perceiving objects that are not present in the image.
Thirdly, for the proposed NNs adversarial samples cause predictions with low confidence, enabling them to be accurately identified. In contrast to humans, our networks still do not correctly classify adversarial samples. However, the behaviour of our networks is a significant step towards human-like behaviour compared to standard DNNs that produce highly confident, but incorrect, classification predictions for adversarial samples.
The increased robustness to adversarial and unknown samples is achieved while maintaining competitive performance in terms of generalisation to novel samples from the training categories.  This enables models of this type to be competitive with state-of-the-art DNNs across multiple different evaluation metrics and benchmark tasks.
\end{abstract}

\paragraph{Code:} \url{https://codeberg.org/mwspratling/STEAL/}


\section{Introduction}
\label{sec-introduction}

Deep neural networks (DNNs) are widely considered to be the best available models of visual cognition \citep{Kubilius_etal19,Lindsay20,YaminsDiCarlo16}.
This is due to the alignment between the neural activity within DNNs and that recorded in visual cortex \citep{Conwell_etal24,Schrimpf_etal20,Khaligh-RazaviKriegeskorte14}, and the human-level performance in tasks such as object recognition and image segmentation.
However, more rigorous and comprehensive evaluations of the behaviour of DNN image classifiers reveal that they exhibit many robustness and generalisation issues not seen in humans \citep{Har-ShalomWeiss25,Geirhos_etal18,Bowers_etal23,Ilievski_etal25,WichmannGeirhos23,Malhotra_etal20,Sa-CoutoWichert21,Geirhos_etal20,Recht_etal18,YadavBottou19,Recht_etal19}.
A particularly prominent issue is that DNNs make highly confident predictions even when they are wrong. This issue is not limited to discriminative DNNs applied to image classification, but seems pervasive in modern AI systems. For example, an analogous issue occurs with generative DNNs, such as large language models, which present false information as fact with high confidence (so called ``hallucinations'') \citep{Zhou_etal24,Marcoccia_etal26}.

DNN models of visual perception are prone to making confidently wrong predictions for samples from both unknown and known classes. Examples of the former include test images from classes that the DNN was not trained on \citep{HendrycksGimpel17,Vaze_etal22}, metamers \citep{Feather_etal23}, and ``fooling images'' that look like pure noise to humans \citep{Nguyen_etal15,Kumano_etal22}. The primary example of the latter are adversarial perturbations which cause the DNN to confidently predict that a sample from one of the training classes comes from a different class \citep{Szegedy_etal14,AkhtarMian18,BiggioRoli18,Goodfellow_etal15,Kurakin_etal17,Eykholt_etal17}. The former behaviour shows that DNNs lack sensitivity to large image differences that for humans are clearly perceivable and which mean that the image contains no recognisable object. While the latter shows that DNNs are simultaneously over-sensitive to small changes which are typically imperceptible or irrelevant to humans.


We hypothesised that these pervasive issues might be a consequence of the design of DNNs which led us to rethink the implementation of some of the fundamental building blocks used by neural networks (NNs). Confidently wrong predictions suggest that the neurons in DNNs are insufficiently selective. Hence, we propose a novel, more selective, integration function (ZINC, \cref{sec-ZINC}). In addition, we introduce a novel pooling layer (SCAM, \cref{sec-SCAM}) that aims to enhance tolerance to class-preserving appearance changes so as to facilitate successful generalisation despite the increased selectivity of the integration layers. We train such networks using a recently introduced loss (HEM, \cref{sec-HEM}). From these re-designed components we build DNNs consisting of alternating integration and pooling layers that we call STEAL (Selectivity and Tolerance Enhanced Alternating Layer) networks. 
  

STEAL Nets have the same structure as early DNNs models of object recognition~\citep{Fukushima80,Fukushima88,RiesenhuberPoggio99b,Serre_etal07,Serre_etal02} which were the fore-runners of convolutional neural networks \citep[CNNs;][]{Krizhevsky_etal12,LeCun_etal98}, however, the standard mechanisms for performing integration and pooling (cross-correlation and max-pooling) are replace by our proposed mechanisms.
In both early CNNs and STEAL Nets, the integration and pooling layers can be interpreted as simplistic models of the simple and complex cells observed in primary visual cortex \citep{HubelWiesel62}. Furthermore, as with simple CNNs, STEAL Nets contain two different types of layer as different mathematical processes are required to generate more selective and more tolerant neural responses \citep{RiesenhuberPoggio99b,Spratling05}.
A more selective representation is produced by a neuron responding to a conjunction of co-active synaptic inputs, whereas a more tolerant representation results from a neuron responding to any one of multiple inputs.
In contrast to existing CNNs, our models are designed to more rapidly build-up selectivity to complex stimuli, while simultaneously more rapidly building-up tolerance to changes in appearance. This allows us to create relatively shallow NNs that have competitive performance to modern very deep DNN architectures. The depth of our networks is more consistent with the depth of the ventral stream, which increases the prospect of having a more direct correspondence between DNN stages and cortical regions. The relative shallowness of our architectures also has relevance to machine learning as hierarchical NNs require the output of one layer to provide the input to the next. Hence, even on parallel hardware, the number of layers defines the minimum number of sequential operations that must be performed. As our networks have relatively few layers they have the potential to improve inference speed and address one concern of frugal AI \citep{Girdhar_etal25}.

Our empirical results (\cref{sec-results}) show that our novel building-blocks succeed in reducing the number of confidently wrong predictions. This leads to enhanced robustness to unknown, out-of-distribution, test samples and to  adversarial attacks. Both these types of test data are classified with very low confidence enabling them to be robustly rejected using a simple threshold on the confidence.
For adversarial samples the behaviour of our networks is still not like humans, but is a huge step in the right direction.
Specifically, our networks are sensitive to adversarial perturbations unlike humans for which such perturbations go unnoticed. 
However, unlike existing DNNs our models do not classify adversarial samples wrongly. Instead, adversarial samples result in low confident predictions as is produced by  other forms of image corruption.
 In other words, for the proposed architecture, in stark contrast to existing DNNs, adversarial samples are not a categorically different type of stimulus.  
Improved robustness is  achieved without sacrificing the ability to generalise to novel samples from known classes. 
This improved performance of our new neural network architecture is also of relevance to work in machine learning where the reliability and trustworthiness of existing models is a concern.


\section{Literature Review}
\label{sec-related_work}
\hide{
Information about object identity is processed by a series of cortical regions along the ventral visual pathway, which begins in primary visual cortex (V1) and ends in inferotemporal (IT) cortex \citep{GoodaleMilner92,FellemanVanEssen91}.
At each stage along this pathway, neurons selectively respond to increasingly complex stimuli, and become increasingly tolerant to stimulus location, scale, and other class-irrelevant changes in appearance.
For example, cells in area V1 are responsive to simple stimuli, such as intensity discontinuities at specific orientations and locations \citep{HubelWiesel62}, while neurons in IT are selective to complex objects, such as faces, appearing anywhere in the visual field \citep{KobatakeTanaka94,RustDiCarlo10,Tovee_etal94,Rolls00,OramPerrett94}.

Different mathematical processes are required to generate more selective and more tolerant neural responses \citep{RiesenhuberPoggio99b,Spratling05}.
A more selective representation is produced by a neuron responding to a conjunction of co-active synaptic inputs, whereas a more tolerant representation results from a neuron responding to any one of multiple inputs.
In deep neural network these two operations are performed by {\bf integration} and {\bf pooling} layers, respectively.
Several early DNNs for object recognition~\citep{Fukushima80,Fukushima88,RiesenhuberPoggio99b,Serre_etal07,Serre_etal02} consisted of alternating integration and pooling layers in order to form more specialised representations in one layer, and more invariant representations in the next.
It was proposed \citep{Fukushima88,RiesenhuberPoggio99b} that these two forms of processing correspond to the functionality of the simple and complex cells observed in primary visual cortex \citep{HubelWiesel62}.


These early DNN architectures were the fore-runners of convolutional neural networks (CNNs) \citep{Krizhevsky_etal12,LeCun_etal98}, a DNN architecture that, following decades of refinement, remains popular in computer vision \citep{He_etal16,Liu_etal22a,TanLe20},
and as an image-computable model of the biological visual system \citep{Linsley_etal25,Conwell_etal24,Schrimpf_etal20}.
}

\subsection{Previous Work on Improving DNN Models of Visual Perception}
\label{sec-bio_plausible_DNNs}

Following the introduction of image-computable models of the biological visual system \citep{Fukushima80,Fukushima88,RiesenhuberPoggio99b,Serre_etal07,Serre_etal02}, the subsequent developments that have produced modern DNN architectures have been driven, primarily, by engineers. As a result no new biologically-inspired principles have been added, and in some respects contemporary models are less biologically-plausible than their predecessors. 
For example, current very deep networks are difficult to reconcile with the rapid object categorisation performed by the much shallower ventral pathway \citep{VanRullenThorpe01,Suzuki_etal23,Kubilius_etal19}.
Furthermore, there has been a reduction in the alignment between the neural activity patterns in brains and the most recent DNN architectures \citep{Linsley_etal23,Linsley_etal25,Nonaka_etal21}.
A number of recent works have advocated for the use of biological-inspiration to improve DNNs \citep{Zador19,Hassabis_etal17,Zhu_etal20,Malhotra_etal20,Linsley_etal25,Celeghin_etal23,Roy_etal24}.
Practical attempts to do this have mainly fallen into two categories: more biologically-plausible learning algorithms, and more ecologically valid training data.

DNNs are trained using back-propagation, which has long been recognised as biologically-impl\-ausible \citep{Crick89b,Lillicrap_etal20}. Over many decades, a substantial body of research has aimed to address this issue by developing alternative learning algorithms that more closely align with biological neural processes while retaining effective training capabilities \citep{Bengio_etal15,Sacramento_etal18b,WhittingtonBogacz17,Lillicrap_etal16,VonHunerbein_etal26}.
Here, we are interested in the performance of a DNN post-optimisation, not the specific method used to perform that optimisation.
Given that more biologically-plausible alternative training procedures typically produce DNNs with worse accuracy, we employ standard back-propagation methods. 

Modifying the training data (\eg data augmentation techniques) have been used extensively to enhance the robustness and generalisation of DNNs. Such techniques have also been used to try to better align DNNs with biology. For example, to mimic the improvement in visual acuity that occurs during infant development, DNNs have been trained with blurred images where the strength of the blurring reduces over the course of training \citep{JangTong21,Sinha_etal20,Raj_etal25}.
Other aspects of human vision that have been emulated through modification to the training data include: the increase colour saturation and contrast sensitivity that occurs during development \citep{Lu_etal25}; the viewing of an object from different viewpoints for an extended period of time  
\citep{Pandey_etal25,Aubret_etal23}; the increased viewpoint diversity that occurs in natural vision compared to typical photographs \citep{Aw_etal25,LuoMuller25}; the effects of gaze-shifts \citep{Schaumloffel_etal25,Yu_etal25a,Thorat_etal25};
and cortical magnification, the uneven representation of visual space across the visual field \citep{Yu_etal25b}.
These approaches are complementary to the methods that we explore in this paper. Hence, they might be used in combination with the architectural modifications we propose, but such a combination is out of scope of the current article.

Rather than modifying the training data, others research has modified the DNN architecture to introduce a pre-processing stage (or new first layer) that modifies the training and test data. Such ``front-ends'' aim to simulate the function of the retina \citep{Stracke_etal25,Kim_etal15}, the lateral geniculate nucleus \citep{Bertoni_etal22}, or primary visual cortex \citep{Evans_etal22,Strisciuglio_etal20,Dapello_etal20}.
The effects of such front-ends could also be produced by using suitable data-augmentations. Hence, these approaches are interchangeable. However, they embody slightly different philosophies. The data augmentation methods use a standard DNN and attempt to influence what it learns by changing its inputs. The front-end methods implement inductive biases into the DNN architecture that will influence what it learns by changing its inputs. The latter approach is therefore closer to the current work in that we seek to modify the computational mechanisms used by a DNN. 
However, in contrast to these existing approaches, rather than seeking specialised modifications for individual layers, the current work aims to define general purpose modifications that can be made to every layer of the same kind (\ie each pooling layer or each integration function). Our approach is consistent with the fact that different cortical regions are remarkably uniform in terms of cortical circuitry and the computations they perform
\citep{Ebdon96,DouglasMartin04,Mumford92b,CrickAsanuma86,DeFelipe97,PhillipsSinger97,Grossberg98aa,Barlow94}.

\subsection{Previous Work on Improving the Response Function}

Both biological and artificial neurons receive multiple inputs and produce a single output. The process of calculating the output is described as a ``response function'' that is typically modelled as a sequence of two operations: the ``integration function''\footnote{This is also commonly referred to as the ``transfer function''. However, ``transfer function'' has been used inconsistently in the literature being used to refer to both the response function and the activation function. To avoid ambiguity we will use the term ``integration function'' here.} that uses the synaptic weights to combine multiple inputs into a single value, and an ``activation function'' that performs a nonlinear operation on the output of the integration function.
The standard integration function calculates a weighted sum of the inputs, through matrix multiplication in fully-connected (FC) layers, or through cross-correlation (CC) in convolutional layers.
The separation of the neural response function into a linear part followed by a nonlinear part is also employed, using different terminology, in rate-based models of biological neurons (``linear-nonlinear'' models), and spiking neural networks (``integrate-and-fire'' models).

While techniques have been proposed to improve the computational efficiency of linear integration \citep{Chollet17,Chen_etal21c,Nowak_etal23,Howard_etal17}, it has remained the standard choice for more than 65 years \citep{Rosenblatt58}. This contrasts with the far larger volume of research on the activation function which  has resulted in a large variety of different activation functions (sigmoid, tanh, ReLU, GELU, ELU, PReLU, SiLU, Mish, Softplus, Softmax, \etc). 
Some architectures, such as LSTM \citep{HochreiterSchmidhuber99b}, transformers \citep{Vaswani_etal17, Graves_etal14}, and some models of cortical pyramidal cells \citep{Phillips_etal24,Spratling02,KordingKonig00a,Adeel_etal23,Georgescu_etal23}, employ multivariate activation functions that operate on the outputs of multiple integration functions. However, in all cases these integration functions combine their inputs using a standard, linear, weighted sum. For example, in vision transformers the three terms used in self-attention (the queries, keys and values) are calculated using FC \citep{Vaswani_etal17,Liu_etal21,Touvron_etal21} or CC \citep{Dosovitskiy_etal20,Han_etal23}.


In contrast to this linear integration performed by artificial neurons, biological neurons integrate their inputs in a highly nonlinear manner, via multiple dendritic compartments that perform complex computational functions \citep{Stuart_etal99,Rall64,Koch_etal83,Mel94,Segev95,SprustonKath04,LondonHausser05,HausserMel03,Spruston08}.    
While work in computational neuroscience has explored the computations performed in the dendrites \citep{Mel93,PoiraziPapoutsi20}, this has usually been restricted to simulating single cells (or small circuits of a few cells) in great detail. With very few exceptions \citep{Xue_etal25,Bertalmio_etal20,SpratlingSchutt25b} nonlinear integration has not been employed in DNNs due to the difficultly in learning the parameters in such integration functions or the very high computational costs that would be involved.
Other work, whether biologically-inspired or not, has used nonlinear integration in only the first layer of a NN \citep{Mel92a,MelKoch90,SpratlingHayes00a,BroomheadLowe88,ParkSandberg91,Azeglio_etal24,SpratlingJohnson01b}.
Some DNNs replace single neurons with small neural networks \citep{Cang_etal24,Lin_etal14b}. This can be interpreted as implementing the idea that nonlinear dendritic computation can be modelled using a multi-layer NN \citep{Poirazi_etal03,Beniaguev_etal21}: \ie as a sequence of linear integration and nonlinear activation functions. If a multilayer NN is required to model of dendritic integration of a single biological neuron, this might explain the current need for very deep NN architectures.

\begin{figure*}[tb]
  \begin{center}
    \hspace*{4mm}\makebox[0.245\textwidth][c]{\small Template}\makebox[0.245\textwidth][c]{\small Search Image}\makebox[0.245\textwidth][c]{\small CC}\makebox[0.245\textwidth][c]{\small ZINC}\\
      \rotatebox{90}{\hspace*{2mm}\small Uneven Intensity}\hfill
    \includegraphics[width=0.97\textwidth,trim=5 5 5 5, clip]{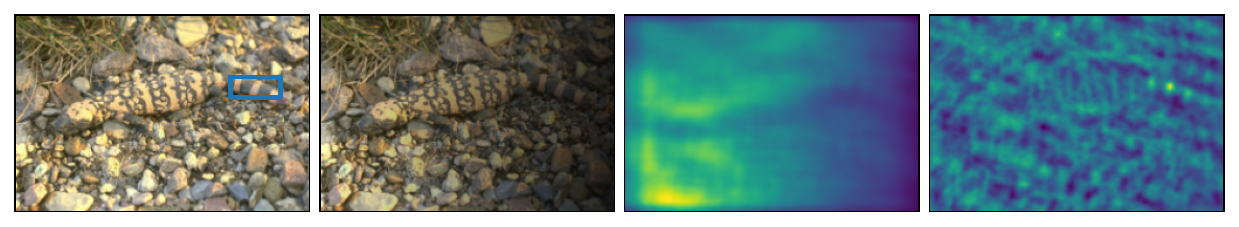} 
      \rotatebox{90}{\hspace*{5mm}\small Blur \& Noise}\hfill
    \includegraphics[width=0.97\textwidth,trim=5 5 5 5, clip]{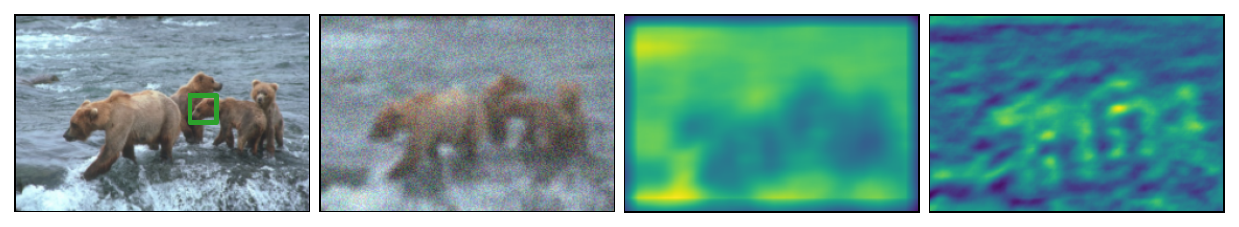} 
      \rotatebox{90}{\hspace*{6mm}\small Unmodified}\hfill
    \includegraphics[width=0.97\textwidth,trim=5 5 5 5, clip]{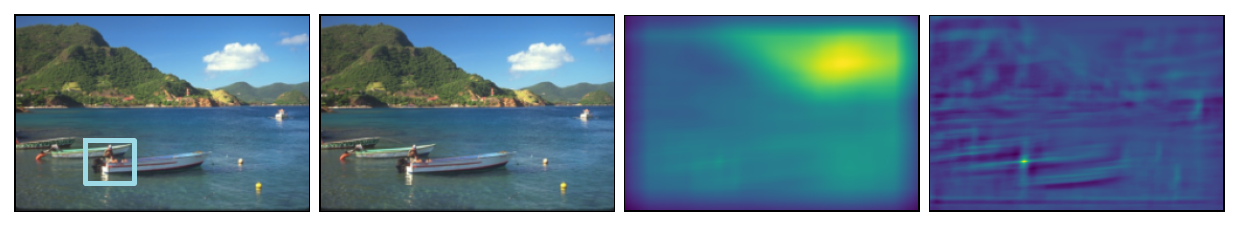} 
    \caption{Comparison of different integration functions on a template matching task. The columns show: 1) the original image and a bounding box indicating the location where the template was extracted; 2) the image to which the template was compared; the similarity between the template and the searched image calculated by
      3) the standard, linear, integration function (CC),
      and 4) the proposed integration function (ZINC). In columns 3 and 4 brighter colors indicate higher similarity. In the first row the search image was produced by non-uniformly changing the intensity of the original. In the second row the search image was produced by blurring followed by the addition of uniform random noise. In the third row the search image is identical to the original. In all cases the highest similarity values produced by the linear integration function are at image locations other than that where the template came from.
      For ZINC the highest output is at the correct location for all three images.
      The images come from the BSDS500 data-set \citep{Martin_etal01}.}
    \label{fig-template_matching}
  \end{center}
\end{figure*}

In this article we implement a computationally tractable nonlinear integration function that is used to replace all integration layers in a DNN. Rather than basing this function on the mechanistic properties of dendrites, it is inspired by the functional properties of biological neurons. Specifically, the characterisation of neurons as ``feature detectors'' \citep{Barlow53}.
Individual neurons are believed to discriminatingly respond only to only a sub-set of all possible stimuli. Much work in neuroscience is predicated on this assumption as this work endeavours to identify the external stimuli to which neurons respond \citep{Usrey_etal03,OramPerrett92,KnierimVanEssen92,Quiroga_etal05,DerringtonLennie84,Wachsmuth_etal94,GawneMartin02,Sharpee_etal04,Rodieck65}.
In contrast, the linear integration function used in existing NNs is poor at selectively responding to specific features (see \cref{fig-template_matching}).
This is not surprising, as it simply calculates a weighted sum of inputs, and so can produce high activity when the input is strong but not necessarily a good match to the pattern of synaptic weights.
As an extreme example, a linear neuron with hundreds of inputs can be driven to respond arbitrarily strongly by a sufficiently large magnitude input to any individual synapse with a non-zero weight. This can occur even when all the other inputs are zero or random.

Work in signal and image processing has attempted to develop more effective feature-detection methods than CC \citep{Spratling_patchmatching}. 
Zero-mean normalised cross correlation (ZNCC, also known as the sample Pearson correlation coefficient) is a classic, and relatively fast, method that is particularly effective when applied to features learnt by a DNN \citep{GaoSpratling21}. We implement an approximate version of ZNCC (called ZINC, see \cref{sec-ZINC}) that can be used as a drop-in replacement for convolution layers in DNNs.

\subsection{Previous Work on Improving the Pooling Function}

Pooling was introduced in early DNN models \citep{Fukushima88,Fukushima80,RiesenhuberPoggio99b,Serre_etal07} 
 to simulate the function of complex cells in visual cortex \citep{HubelWiesel62}. Complex cells are believed to combine the outputs of neurons with
 different selectivities (for example, to produce  responses that are tolerant to contrast polarity), and different receptive field locations (to produce tolerance to spatial location). Biologically-inspired pooling, therefore, has three objectives:
1) to combine responses across channels (\ie selectivities);
2) to combine responses across space; and
3) to reduce the spatial dimensions of a feature-map.
However, pooling functions that are typically used in DNNs (such as max or average pooling) only achieve the last two objectives. Strided convolution, which replaces pooling in many modern DNN architectures, only achieves the last objective.
Hence, while complex cells are considered essential for increasing tolerance to appearance changes in the biological visual system, this functionality has never been fully implemented in DNNs. 
We believe that a lack of spatial pooling results in worse tolerance to changes in spatial location, and lack of channel-wise pooling results in worse tolerance to other class-preserving changes in appearance.
The only mechanisms current CNNs have available to learn these tolerances is through the response function, which is another possible reason for the need for very deep architectures.

Improving the ability of spatial pooling to increase tolerance to location shifts has been an active area of research \citep{Grabinski_etal22,Vasconcelos_etal21,Stergiou_etal23,Zhang19,NingSpratling23,Stergiou_etal21,Li_etal21,Zou_etal20}.
In contrast, apart from our own preliminary work \citep{SpratlingSchutt25b} the only channel-wise pooling function, Channel Max Pooling \citep{Cheng_etal19}, implements channel-wise pooling separately from spatial pooling as a specialised operation that is only performed once after the last convolutional layer in a CNN. 
In this article we implement a pooling function (called SCAM, see \cref{sec-SCAM}) that simultaneously operates across space and a sub-set of channels and that is used in every block of a DNN.

\section{Methods}
\label{sec-methods}

\subsection{The ZINC Integration Function}
\label{sec-ZINC}

The standard, linear, integration function is defined as: $y_{CC}=\mathbf{w}^T\mathbf{x} = \sum_i w_ix_i$, where $\mathbf{x}$ is a column vector of inputs, $\mathbf{w}$ is a column vector of synaptic weights, and $y$ is the scalar output of the integration function.
ZNCC is implemented as:
\begin{equation}
  y_{ZNCC}=\frac{\sum_i (w_i-\bar{w})(x_i-\bar{x})}{\sqrt{\sum_i (w_i-\bar{w})^2\sum_i (x_i-\bar{x})^2}}
  \label{eq-ZNCC}
\end{equation}
where $\bar{w}$ and $\bar{x}$ are the mean values of the vectors $\mathbf{w}$ and $\mathbf{x}$, respectively. Implementing ZNCC in a convolutional-like manner would require $x_i-\bar{x}$ to be separately calculated for each sliding window, as the same element in the tensor representing the inputs to a layer will appear in multiple windows each with a different mean value. To avoid the computational overhead this would entail, we approximate ZNCC by simply subtracting an estimate of the local mean from the input tensor (see \cref{alg-ZINC}). As a result in our approximate implementation, $\bar{x}$ denotes the mean activity of a region of the input tensor rather than the mean of the specific window used to define the vector $\mathbf{x}$. The size of the spatial window used to calculate the mean of the input, $\zeta$, is set to be $\lambda$ times the convolution kernel size (rounded up to the next highest odd integer), where $\lambda$ is a hyper-parameter that was set to a default value of two. Due to this approximation, we call our integration function Zero-mean (Imprecise) Normalised Cross-correlation (ZINC). 

\begin{algorithm}[tbp]
  \caption{Pseudo-code for ZINC implemented as a pytorch-style replacement for standard 2d convolution.}
  \label{alg-ZINC}
\begin{minted}{python}
def ZINC2d(x, w, beta, zeta, stride=1, padding=0):
    """
    x       a tensor containing the inputs, with shape (B, Ci, H, W)
    w       a tensor defining the weights, with shape (Co, Ci, k, k)
    beta    a non-negative, scalar, used to prevent division-by-zero errors and
            define the smallest magnitude input considered to be significant
    zeta    the kernel size used to calculate the local spatial mean of x
    stride  an integer defining the stride of the operation
    padding an integer defining the amount of padding applied to the input
    """
    # remove a local estimate of the mean from x
    x = x - avg_pool2d(mean(x, dim=1, keepdim=True), kernel_size=zeta, 
                       stride=1, padding=zeta//2, count_include_pad=False)
    # calculate the squared l2-norm of the weights
    wNorm = sum(w**2, dim=(1,2,3))
    # calculate the squared l2-norm of the inputs}
    xNorm = avg_pool2d(sum(x**2, dim=1, keepdim=True), kernel_size=k, 
                       stride=stride, padding=padding) * k**2
    # calculate the normalised cross-correlation between the zero-mean inputs
    # and the weights
    y = conv2d(x, w) / sqrt( wNorm[None,:,None,None] * (xNorm+beta) )
    return y
\end{minted}
\end{algorithm}

In ZINC, the weights, $\mathbf{w}$, are initialised with random values from a normal distribution with a mean of zero, and an $l_2$-norm of one, and are then optimised using back-propagation. We do not explicitly enforce that the learnt weights have zero-mean, however, back-propagation tends to learn weights such that the mean is close to zero. Some deviation from zero is expected, and desirable, as the imprecise operation used to remove the mean from the inputs (as discussed in the previous paragraph) will cause the input to each sliding window to not exactly be zero-mean.
Learning maintains the $l_2$-norm of the weights to be considerably greater than zero. However, it is possible for the inputs to a ZINC layer to contain regions where the $l_2$-norm of the (approximately zero-mean) inputs is zero. Hence, to prevent division-by-zero errors and to define the smallest magnitude input considered to be significant, $\beta$, a non-negative scalar is added to the squared $l_2$-norm in the denominator of the integration function. Hence, ZINC is defined as:
\begin{equation}
  y_{ZINC}=\frac{\sum_i w_i(x_i-\bar{x})}{\sqrt{(\sum_i w_i^2)(\sum_i (x_i-\bar{x})^2 + \beta)}}
  \label{eq-ZINC}
\end{equation}
where $\bar{x}$ is the local mean values in the input tensor, and $\beta$ is a non-negative scalar shared by all the neurons in a layer.
The value of $\beta$  was made equal to a running estimate the mean value of $\gamma\sum_i (x_i-\bar{x})^2$, where  $\gamma$ is a hyper-parameter that was set to a default value of 0.1. This estimate was updated, using a learning rate of 0.01, during the forward pass when training, analogously to the way statistics are updated in batchnorm layers \citep{IoffeSzegedy15}. 
ZINC is far better at feature-detection than the standard, linear, integration function, as illustrated in \cref{fig-template_matching} .


\subsection{The SCAM Pooling Function}
\label{sec-SCAM}

For the more selective neurons, it is important to enable the network to learn to generalise to changes in appearance by allowing pooling over multiple channels. Thus, pooling was performed over a number of input channels in additional to a spatial region. Each output channel combined values from a disjoint set of $\kappa$ input channels. The set of input channels in each pool was pre-defined and fixed. We implement our channel and spatial pooling using the $avg\_pool3d$ function in pytorch (see \cref{alg-SCAM}).

All values in the 3-dimensional pooling region, $\mathds{P}$, within the input feature map, $\mathbf{y}$, were combined using the Mellowmax operator \citep{AsadiLittman17}:
\begin{equation}z= \frac{1}{\alpha}\log\left[ \frac{1}{\vert\mathds{P}\vert} \sum_{i \in \mathds{P}}\exp(\alpha y_i)\right]\label{eq-SCMP}\end{equation}
Depending on the value of $\alpha$ this function can produce a smooth approximation to min pooling ($\alpha\ll 0$), average pooling ($\alpha\approx0$), or max pooling ($\alpha\gg 0$). A single value of $\alpha$ was used by each layer, and its value was learnt through back-propagation.
$\alpha$ was initialised to a value of one but the final, learnt, values tended to be large. The gradual transition from an approximation of average pooling towards a smooth version of max pooling was found to be beneficial. This may be because it ensures that all channels in the pool receive strong gradients early in training, and hence, all channels learn. While later in training max-like pooling can encourage different input channels to learn distinct selectivities. The value of $\alpha$ learnt by different layers was observed to often be very different. Hence, learning also enabled the style of pooling to be adapted to different layers in the DNN hierarchy. We call this pooling function Spatial and Channel-wise Adaptive Mellowmax (SCAM).




\begin{algorithm}[tbp]
  \caption{Pseudo-code for SCAM implemented as a pytorch-style replacement for standard 2d pooling.}
  \label{alg-SCAM}
\begin{minted}{python}
def SCAM2d(y, k, kappa, alpha, stride=k):
    """
    y      a tensor containing the inputs, with shape (B, Ci, H, W)
    k      an integer defining the spatial size of each pool
    kappa  an integer defining the number of input channels in each pool
    alpha  a scalar defining how mellowmax combines values
    stride an integer defining the spatial stride of the pooling operation
    """
    # apply the mellowmax operator to each pool region
    z = log(avg_pool3d(torch.exp(alpha * y), (kappa, k, k), 
                       stride=(kappa, stride, stride))) / alpha            
    return z
\end{minted}
\end{algorithm}

\subsection{The HEM Loss Function}
\label{sec-HEM}

We found that standard cross-entropy (CE) loss was poor at training networks built using ZINC and SCAM, and that better results were achieved using High Error Margin (HEM) loss \citep{SpratlingSchuttHEM} a recently proposed variant of multi-class margin (MM) loss \citep{CrammerSinger02}. For a vector of network outputs  $\mathbf{z} \in \mathds{R}^n$, where each output corresponds to one of $n$ classes, margin losses compute an error $e_i$ for each output, $z_i$:
\begin{equation}
e_{i \ne l}= \max(0,z_i-z_l+\mu) \;\;\;\;\;\;\;\;\;\; e_l=0
\label{eq-MM_error}
\end{equation}
Where $l$ is the index corresponding to the correct class, and $\mu$ is a non-negative hyper-parameter defining the margin. 
While MM loss combines the errors for different outputs (and across all samples in a batch) by taking the mean, HEM loss averages only the errors which are larger than the average error:
\begin{equation}
  \mathcal{L}_{HEM} = \langle\mathbf{E}\rangle \;\;\; \text{where} \;\;\;\mathbf{E} = \{ e \;\vert\; e > \langle e \rangle \}
  \label{eq-HEM}
\end{equation}
The losses for different samples in the batch are combined by finding the mean of above-zero values.

HEM loss has been found to improve the robustness of standard CNNs (particularly, the ability to reject adversarially perturbed samples, and samples from unknown classes), but this improvement is sometimes at the expense of poorer generalisation performance (lower accuracy with clean and corrupted test images) \citep{SpratlingSchuttHEM}. Here, we find that combining HEM with ZINC and SCAM over-comes this trade-off to simultaneously improve performance on all these metrics.

\subsection{Experimental Set-up}
\label{sec-expt_setup}

\subsubsection{Neural Network Architectures}
\label{sec-NN_archs}

To evaluate the proposed new building-blocks they were used in fully-convolutional-like NN architectures consisting of repeating blocks, each containing an integration function followed by an activation function followed by pooling. The number (B) of these blocks, the width and height of the integration function's square receptive-field (W), and the number of output channels (N) in each block were varied to produce networks of different sizes appropriate for different training data-sets. The structure of a particular network is described using the short-hand: B-W-N.
For ImageNet1k it is desirable to use a large number of channels to reflect the large number of classes. However, in early blocks where the width and height of the feature map is large, due to the resolution of the images, using a large number of channels is very computationally demanding. For this data-set we therefore allow the number of channels to vary with depth. Specifically, N defines the number of output channels from the first block, and the number of channels in each subsequent block is increases by a multiplicative factor, I, compared to the preceding block. I is made equal to $(\frac{n}{N})^{(\frac{1}{B-1})}$, where $n$ is the number of channels in the last block, \ie the number of classes. The structure of such networks is denoted as B-W-N-exp. Many modern DNNs similarly increase the number of channels in deeper blocks.

For networks composed of standard components each block consisted of convolution, ReLU, and max pooling. For networks built from our proposed components each block consisted of ZINC (\cref{sec-ZINC}), ReLU, and SCAM (\cref{sec-SCAM}). In both cases, the activation function, ReLU, was removed from the final block for compatibility with CE loss.
For  networks built from the proposed components the number of input channels, $\kappa$, in each pooling region was the same in each block. The spatial size of the pooling regions was identical for both standard CNNs and STEAL Nets and was between 1 and 4. Padding of either 0 or 1, together with the spatial pool size, was used to create valid architecture for the image sizes used. 
We compared performance for analogous architectures composed of standard and the proposed building-blocks. The compared networks had the same depth, kernel dimensions, spatial pooling size, and padding. They differed in terms of the number of output channels in each block, N, which was adjusted make the number of trainable parameters as similar as possible in the compared networks. N was smaller for the networks built from the proposed building-blocks
because there are N$\times \kappa$ ZINC layer output channels (due to the channel-wise component of SCAM).


\subsubsection{Training Settings}
\label{sec-training_setup}

The standard networks were trained using CE loss, while the proposed models were trained using HEM loss with the margin set to $\mu = 0.1$ (\cref{sec-HEM}).
The AdamW optimiser \citep{LoshchilovHutter19} and the One-Cycle learning schedule \citep{SmithTopin18} were used. These were both chosen to help ensure stable learning without the need to perform a hyper-parameter search.
  To ensure that this was a reasonable choice that did not favour either architecture we evaluated a number of combinations of optimiser and scheduler, and results are reported in \cref{sec-results_training_setup}.
A learning rate of 0.01 was found to lead to successful training in the proposed architectures, except when training with ImageNet1k where a learning rate of 0.001 was used. For the standard architectures the best results on each benchmark produced using learning rates of either 0.01 or 0.001 are reported.
A weight decay of 0.04 was used as this was found to produce slightly better results for the standard networks than using  the pytorch default weight decay for AdamW (0.01). Note, that weight decay has little effect on ZINC due to the normalisation of the activity by the $l_2$-norm of the weights.

10, 20, 20, 110, 110, 200, and 100 training epochs were used for MNIST \citep{LeCun_etal98}, FashionMNIST \citep[FMNIST;][]{Xiao_etal17}, SVHN \citep{Netzer_etal11}, CIFAR10 \citep{Krizhevsky09}, CIFAR100 \citep{Krizhevsky09}, TinyImageNet (TinyIN), and ImageNet1k \citep{Russakovsky_etal15}, respectively. 
The batch size was 128 for all data-sets. Pixel values in both the training and testing samples were scaled to the range [0,1]. For all data-sets the standard split between training and testing exemplars was employed.
For each data-set standard, basic, data augmentations were applied to the training images: no augmentation for MNIST and SVHN, random horizontal and vertical flipping for FashionMNIST, and random horizontal flipping and cropping for the others. 

For some experiments (\cref{sec-results_robustbench}) a stronger training data augmentation scheme was used. This consisted of the basic augmentations followed by a sequence of seven additional augmentations, applied in random order, each with a probability 0.2 of being applied, and each with a random strength. Three of these augmentations modified the lighting conditions by changing the saturation, contrast and brightness. Two of these augmentations removed information either by applying a Gaussian blur or random erase \citep{Zhong_etal17}. Three of these augmentations introduced noise by randomising the magnitude, in the Fourier domain, or mixing the image with random pixel values, or adding sparse Gaussian noise. 
  Following these strong augmentations the Fast Gradient Sign Method \citep[FGSM;][]{Goodfellow_etal15} was applied, also with a probability of 0.2, to produce adversarial perturbations in 20\% of the augmented images. This very weak form of adversarial training (AT) was replaced by a stronger method, seven-step Projected Gradient Descent \citep[PGD;][]{Madry_etal18} applied to 100\% of the images 
  for some of the experiments described in \cref{sec-results_AT}.
  

\subsubsection{Performance Metrics}\label{sec-metrics}

\begin{figure}[tbp]
\begin{center}
\includegraphics[width=0.2\textwidth]{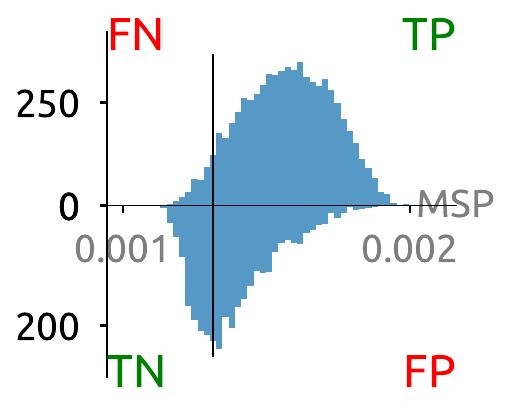}
\caption{Confidence histogram format. This figure shows (on the horizontal axis) the range of confidence scores produced by a classifier and  (on the vertical axis) the number of samples classified with that confidence. Confidence scores are measured using Maximum Softmax Probability \citep[MSP;][]{HendrycksGimpel17}. The histogram above the horizontal axis shows the distribution of confidence scores associated with samples for which the classifier has predicted the correct class label. Note for novel and unrecognisable test data, which come from unknown classes, this histogram is blank. The inverted histogram below the horizontal axis shows the distribution of confidence scores associated with samples for which the classifier has predicted the wrong class label. The black vertical line shows the threshold used to reject or accept the classifier's prediction (this threshold is set so that 95\% of correctly classified clean examples were accepted). In the upper histogram samples to the left of this threshold are false-negatives (samples rejected due to low confidence despite having been assigned the correct label), while samples to the right of this threshold are true-positives (samples correctly classified with high confidence). In the lower histogram samples to the left of the threshold are true-negatives (samples correctly rejected because the assigned label was incorrect), while samples to the right of the threshold are false-positives (samples incorrectly classified with high confidence).}
\label{fig-MSP_hist_truth_key}
\end{center}
\end{figure}

The methods described in \citet{Spratling_robustnessevaluation} were used to quantify performance. This benchmark rigorously and comprehensively tests the performance of a DNN across a range of different data-types: the standard test data (``Clean'' data); the common corruptions data (``Corrupt'' data); adversarially-perturbed images (``Advers.'' data); images of objects from unknown classes (``Novel'' data); and, synthetic, noise, images containing no object (``Unrecog.'' data). Humans perform accurately on each of these types of data, hence, we equate better results with improved alignment between the model and human perception. Performance on each data-type is assessed, consistently, using the same metric: detection accuracy rate (DAR). DAR measures the proportion of true-positives (samples correctly classified with high confidence) and true-negatives (samples given low confidence and wrong class labels). The threshold separating high and low confidence was set such that 95\% of correctly classified clean examples were accepted.
The confidence score was determined using Maximum Softmax Probability \citep[MSP;][]{HendrycksGimpel17} which uses the maximum value of the network output after the application of the softmax activation function. However, the results were not dependent on this choice as similar results were obtained using 
Maximum Logit Score \citep[MLS;][]{Vaze_etal22,Hendrycks_etal22b}, which defines the confidence as the maximum, raw, response of the final layer. 
Confidence histograms, like that shown in \cref{fig-MSP_hist_truth_key}, are plotted to provide a  more in-depth analysis in some experiments. The format of this type of figure is described in the caption of  \cref{fig-MSP_hist_truth_key}.

The exact make-up of each data-type used in this evaluation is described below.  
\begin{description}
\item[Clean.] The standard test-set provided with each data-set.
\item[Corrupt.] Common corruptions data-sets: MNIST-C, FashionMNIST-C (FMNIST-C), CIFAR10-C, CIFAR100-C, Tiny\-ImageNet-C and ImageNet-C \citep{HendrycksDietterich19,MuGilmer19,WeissTonella22}. These data-sets modify the standard test-set using multiple corruption methods including  
  different types of noise, blurring, geometric transformations, superimposed patterns, synthetic weather conditions, and digital corruptions.
  As typical in the literature, performance is evaluated using all the corruptions at all degrees of intensity.
When the test-set resolution differs from that of the training set, the test images is resized (using bi-linear interpolation) to match the image resolution used for training. Similarly, the number of colour channels in the test images is modified, if necessary, to match the training data (\eg grayscale images are converted to colour by making the three colour channels all equal to the original image intensity values).
\item[Adversarial.] Samples are produced using AutoAttack \citep[AA;][]{CroceHein20}, a state-of-the-art ensemble attack method that employs both gradient-based (white-box) and gradient-free (black-box) attacks.
  Two sets of adversarial samples are used. Each set is created by perturbing the entire standard (clean) test-set\footnote{Do to the extremely high computational demands of performing AA on large ImageNet1k trained networks, only 1000 adversarial samples were used to test these networks.}, but with a different method of constraining the magnitude of the perturbation. Specifically, AA is used to apply both $l_\infty$ and $l_2$-norm constrained attacks.
  The perturbation budget ($\epsilon$) used for each attack is the standard value used in the previous literature for each data-set. Specifically, $\epsilon$ for $l_\infty$-norm and $l_2$-norm) attacks was set to 0.3 and 2 for networks trained on MNIST and FMNIST, $\frac{4}{255}$ and 0.5 for ImageNet1k-trained networks, and $\frac{8}{255}$ and 0.5 for networks trained with all other training data-sets. 
\item[Novel.] Three data-sets containing objects of unknown classes are used for each training data-set. Specifically,
  MNIST trained networks are tested using unknown, novel, classes from the test-sets of the Omniglot \citep{Lake_etal15}, FashionMNIST \citep{Xiao_etal17} and KMNIST \citep{Clanuwat_etal18} data-sets.
FashionMNIST trained networks are evaluated with the test-sets of the Textures \citep{Cimpoi_etal14}, MNIST and CIFAR10 data-sets. When trained with SVHN, unknown class rejection is evaluated with the Textures, Icons-50 \citep{HendrycksDietterich19}, and CIFAR10 test data-sets.
  The CIFAR10 and CIFAR100 trained networks are tested using novel classes from the test-sets of the Textures and SVHN data-sets and the CIFAR data-set that was not used for training (\ie a CIFAR10 trained network is evaluated on its ability to reject samples from CIFAR100, and vice versa).
  For TinyImageNet and ImageNet1k trained networks novel class data comes from the Textures test set, the iNaturalist 2021 validation set \citep{VanHorn_etal18}, and the ImageNet-O data-set \citep{Hendrycks_etal19c}.
When the test-set resolution differs from that of the training set, the test images is resized (using bi-linear interpolation) to match the image resolution used for training. Similarly, the number of colour channels in the test images is modified, if necessary, to match the training data (\eg colour images are converted to grayscale).

\item[Unrecognisable.] Data to test unrecognisable image rejection is the same for all training data-sets, and consisted of synthetic images randomly generated using four different methods.
  First, the images of the standard (clean) test set after randomising the phase, in the Fourier domain, of each image (``Phase'').
  Second, the images of the clean test set after a random permutation of all pixels (``Scramble''). 
  Third, as used in \citet{Hendrycks_etal19}, images containing random blobs (``Blobs'').
  Fourth, images in which each pixel intensity value is independently and randomly selected from a uniform distribution (``Uniform''). 
  Each of these four data-sets contain the same number of samples as the standard test-set associated with the training data.
\end{description}

Traditional evaluation metrics are also reported in the appendix. Specifically, the percentage of samples correctly classified from the clean data, the corrupt data, and the adversarial data. The 
Area Under the Receiver Operating Characteristic curve (AUROC) is reported to assess out-of-distribution (OOD) rejection performance for the novel and unrecognisable data combined.
AUROC is calculated separately for each unknown class data-set, evaluating how distinct the MSP confidence scores produced by samples from the standard test-set are from the confidence scores produced in response to samples from the unknown class data-set. It should be noted that these traditional metrics are incompatible with each other and do not provide a true indication of realistic performance in the real-world \citep{Spratling_robustnessevaluation}. For example, AUROC is calculated using separate, optimal, rejection thresholds for each OOD data-set: so the performance on one data-set is incompatible with another when the required thresholds differ.  Furthermore, no rejection threshold is applied when assessing performance with in-distribution test data-sets. Hence, using these traditional metrics a network might appear capable of classifying corrupt data, but if its prediction confidence for this data is similar to or lower than that produced for OOD data, then the performance on the corrupt data is unachievable simultaneously with the reported performance on the OOD data.

For all experimental conditions (except those using ImageNet1k) a network was trained and evaluated five times,
each time with a different random weight initialisation and random presentation order of training samples. The mean and standard-deviation over these trials is reported, and the two-sample t-test (with $p<0.05$) was used to determine if differences in performance were statistically significant.

\subsection{Code}

Open-source code implemented using Python using the PyTorch library \citep{Pytorch19} is available from: \url{https://codeberg.org/mwspratling/STEAL/}.

\section{Results}
\label{sec-results}

We assume that good models of visual perception should be able to perform simple visual tasks such as: recognising novel exemplars of known objects (even when this requires generalisation to situations unseen in training); not being fooled by adversarial attacks; and not mistaking unknown, novel, objects or noise for known objects. We therefore test STEAL networks on such tasks and show that they perform better than DNNs built from standard components (\cref{sec-results_size_matched,sec-results_various_archs,sec-results_robustbench}).
The subsequent analysis in \cref{sec-results_confidence,sec-results_adversarial,sec-results_AT} explores how STEAL nets differ from standard CNNs, particularly when under adversarial attack where the differences are greatest. The remaining results, \cref{sec-results_learning,sec-results_training_setup,sec-results_sensitivity,sec-results_ablation}, provide checks on the rigour of our experimental procedures and the generality of the results.

\subsection{Comparison to Size-Matched Networks with the Same Architecture}
\label{sec-results_size_matched}

To allow a fair comparison between our proposed building-blocks and the standard ones, we used very simple NNs unembellished by any potential confounding factors (see \cref{sec-NN_archs}). For the same reason we used very basic training data augmentations (see \cref{sec-training_setup}), rather than state-of-the-art schemes that have been extensively optimised to enhance performance of standard architectures. We compared the performance of STEAL networks composed of the proposed building-blocks to size-matched CNNs with the same architecture built from standard components (see \cref{sec-NN_archs}), on seven standard image classification benchmarks (see \cref{sec-training_setup}). Performance was assessed using many different test-data sets from five different categories of data-type (see \cref{sec-metrics}). Over the 35 different combinations of task and test-data type, the standard networks were superior in only five conditions (and in only two was this difference statistically-significant), while the proposed components led to superior performance in 30 combinations (\cref{tab-results_DAR_all}). In many of these cases, the improvement in performance due to the use of the proposed components is very large, and statistically-significant. This superior performance is not only captured by the DAR metric, but also by traditional evaluation methods (\cref{tab-results_conventional_all}).

In general, the proposed networks reduce both under-generalisation (as shown by improved performance with clean and corrupt data) and over-generalisation (as shown by improved performance with novel and unrecognisable data). This enhances the alignment between the proposed models and the behaviour of the biological visual system as well as increasing the reliability of CNNs for machine learning applications.  

\begin{table*}[tbp]\centering
  \caption{A comparison of the performance of STEAL networks composed of
    the proposed building-blocks (ZINC, SCAM, and HEM)
    with size-matched CNNs composed of standard building-blocks (CC, Max pooling, and CE).
    Bold text indicates the best performance on each metric for each training data-set. Coloured text indicates that the difference in performance is statistically significant.
  }
  \label{tab-results_DAR_all}
    \tabcolsep2pt
    \resizebox{\ifdim\width>\linewidth\linewidth\else\width\fi}{!}{%
      \begin{tabular}{llllllll ccccc *{1}{S[table-format=2.2(2)]}}\toprule
&Integrat.  &Pooling &Loss & Data-set & Network        &Num.    & Training & Clean &	Corrupt	& Advers. &	Novel	& Unrecog. & {Mean $\pm$ Std}\\
&Func.      &Func.   &Func.&          & Architecture   &Params. & Time (s) & DAR (\%) &DAR (\%) &DAR (\%) &DAR (\%) &DAR (\%) &{DAR (\%)}\\  \midrule
our& ZINC   & SCAM   &HEM  & \multirow{2}{*}{MNIST}    & 2-5-20  & 44.0k  & 56.45   &  94.94&	69.30&\W89.26&\W94.01&\W99.73&\W89.45	(0.21)\\
std& CC     & Max    &CE   &                           & 2-5-160 & 44.2k  & 54.02   &\B94.96&\B	69.92&	1.36 &  83.50&	80.91&	66.13	(1.84)\\\midrule
our& ZINC   & SCAM   &HEM  & \multirow{2}{*}{FMNIST}   & 3-5-16  & 86.4k  & 160.02  &  90.61&\W	67.79&\W53.42&\W93.19&\W99.49&\W80.90	(0.54)\\
std& CC     & Max    &CE   &                           & 3-5-54  & 87.9k  & 157.75  &\B90.67&	62.83&	4.29 &  52.64&	41.67&	50.42	(2.63)\\\midrule
our& ZINC   & SCAM   &HEM  & \multirow{2}{*}{SVHN}     & 3-5-30  & 258k   & 233.51  &\W93.51&\W	71.36&\W64.28&\W90.64&\W96.24&\W83.21	(0.61)\\
std& CC     & Max    &CE   &                           & 3-5-96  & 262k   & 217.07  &  90.57&	57.77&	16.39&	40.49&	63.30&	53.70	(0.65)\\\midrule
our& ZINC   & SCAM   &HEM  & \multirow{2}{*}{CIFAR10}  & 5-3-44  &459k    & 2209.3  &\W91.50&\W	80.26&\W69.85&\W66.32&\W92.25&\W80.04	(1.17)\\
std& CC     & Max    &CE   &                           & 5-3-128 &458k    & 1890.5  &  89.49&	77.47&	5.01 &  47.25&	34.59&	50.76	(1.57)\\\midrule
our& ZINC   & SCAM   &HEM  & \multirow{2}{*}{CIFAR100} & 5-3-41  &667k    & 2072.6  &\W76.34&\W	62.75&\W54.78&	30.61&\W32.30&\W51.36	(1.06)\\
std& CC     & Max    &CE   &                           & 5-3-142 &677k    & 1913.1  &  71.86&	58.02&	11.41&\F34.95&	18.11&	38.87	(0.59)\\\midrule
our& ZINC   & SCAM   &HEM  & \multirow{2}{*}{TinyIN}   & 7-3-60  &2.17M   & 39664   &\W63.82&\W  46.54&\W41.00&\B25.45&	 6.15&\W36.59	(1.05)\\
std& CC     & Max    &CE   &                           & 7-3-200 &2.17M   & 24090   &  53.56&	36.90&	14.96&	25.15&\F14.45&	29.00	(0.53)\\\midrule
our& ZINC   & SCAM   &HEM  & \multirow{2}{*}{ImageNet1k}&9-3-15-exp  &27.6M& 142908  &\B75.80&\B	72.71&\B78.20&\B44.15&\B67.65&\B67.70\\
std& CC     & Max    &CE   &                            &9-3-425-exp &27.2M& 196850  &  72.77&    71.03&	13.65&	43.96&	67.08&	53.70\\\midrule
\multicolumn{8}{r}{performance gap (our-std) averaged over data-sets} &3.23   &5.25 &54.82   &16.63 &24.81  &+20.95  \\
\bottomrule \end{tabular}}\end{table*}

\subsection{Comparison to DNNs of Various Architectures}
\label{sec-results_various_archs}

The previous section compared networks with the same, basic, architecture. One objection to this comparison might be that this basic architecture when composed of standard building-blocks represents a weak baseline, as it does not include any of the multiple refinements and improvements that have been made to DNNs over the last few decades. To address this concern, we compare the performance of the basic architecture composed of the proposed building-blocks (\ie a STEAL Net) to a range of well-known DNNs. We include in this comparison existing shallow architectures (AlexNet \citep{Krizhevsky_etal12} and VGG \citep{Simonyan_etal15}), ResNets \citep{He_etal16}, and representative examples of state-of-the-art (SOTA) CNN and transformer based architectures (ConvNeXt \citep{Liu_etal22a} and ViT \citep{Dosovitskiy_etal20}). When these networks are trained on ImageNet1k under the same conditions used to train our network (basic training data augmentations and 100 training epochs), our model outperforms all of them in terms of Mean DAR (\cref{tab-results_DAR_ImageNet_basic}). The STEAL network built from ZINC and SCAM layers has performance on clean, corrupt, novel, and unrecognisable data that is near the middle of the range produced by the other architectures, however, it has outstanding performance with adversarial images (as can also be seen in \cref{tab-results_DAR_all} for other data-sets). The proposed architecture is, therefore, an outlier in terms of it susceptibility to adversarial attacks and this will be investigated in more depth in \cref{sec-results_confidence,sec-results_adversarial}.

These results demonstrate that the proposed building-blocks produce DNN architectures that are competitive with existing ones, even SOTA models that contain many more layers. Such shallow architectures have a better alignment with the structure of the biological visual system as well as having the potential to provide lower-latency inference times relevant to many applications in edge and frugal AI. 

\begin{table*}[tbp]\centering
  \caption{Comparison of different DNN architectures trained on ImageNet1k for 100 epochs with basic training data augmentations. The first row shows result is for a network composed of the proposed components (ZINC, SCAM, and HEM). These results also appear in \cref{tab-results_DAR_all} and are repeated here for convenience. The remaining rows shows results for a variety of DNN architectures, ordered by increasing number of integration function layers.}
    \label{tab-results_DAR_ImageNet_basic}
    \tabcolsep2pt
    \resizebox{\ifdim\width>\linewidth\linewidth\else\width\fi}{!}{%
      \begin{tabular}{llllllll ccccc c}\toprule
 &Integrat.&   Pooling  &Loss       &Network                &Num.     &Time Per &Num. Int.  & Clean &	Corrupt	& Advers. &	Novel	& Unrecog. & Mean\\
 &Func.    &   Func.    &Func.      &Architecture           &Params.  &Epoch (s)&Layers     & DAR (\%) &DAR (\%) &DAR (\%) &DAR (\%) &DAR (\%) &DAR (\%)\\  \midrule
our&ZINC   &   SCAM     &HEM        &9-3-15-exp      &27.6M    & 1429.1 &$9$   &  75.8&   72.7&\B78.2&  44.2&  67.7&\B67.70\\
   &CC/FC  &   Max      &CE         &AlexNet                &61.1M    & 1288.2 &$8$   &  67.3&	62.4&	15.5&	38.0&	55.0&	47.66\\
   &CC/FC  &   Max      &CE         &VGG11                  &133M     & 1798.8 &$11$  &  77.7&	75.3&	10.4&	48.3&	73.8&	57.10\\
   &CC/FC  &   Max      &CE         &VGG16                  &138M     & 1604.0 &$16$  &  79.3&\B	76.5&	10.5&\B	51.2&	67.3&	56.96\\
   &CC/FC  &   Strided  &CE         &ResNet18               &11.7M    & 1268.2 &$21$  &  78.0&    75.0&	12.7&	49.2&\B	87.2&	60.44\\
   &CC/FC  &   Strided  &CE         &ResNet50               &25.6M    & 1387.7 &$54$  &\B81.2&	74.8&	9.27&   48.5&	32.3&	49.19 \\
   &CC/FC  &   Strided  &CE         &ConvNeXt-tiny          &28.6M    & 1109.4 &$59$  &  80.0&	73.7&	10.0&	46.9&	73.0&	56.73\\
   &FC/CC  &   None     &CE         &ViT-S16	            &22.1M    & 1101.4 &$74$  &  77.2&	71.3&	12.8&	46.4&	73.2&	56.17\\
   &FC/CC  &   None     &CE         &ViT-B16	            &86.6M    & 2505.4 &$74$  &  76.3&	70.7&	14.6&	46.8&	64.4&	54.57\\
   &CC/FC  &   Strided  &CE         &ConvNeXt-base          &88.6M    & 1778.9 &$113$ &  80.9&	74.6&	10.7&	48.2&	55.6&	53.99\\
\bottomrule\end{tabular}}\end{table*}

\newcolumntype{C}[1]{>{\centering\let\newline\\\arraybackslash\hspace{0pt}}m{#1}}
\begin{table}[tbp]
\begin{center}
  \caption{Comparison to models ranked highly on the RobustBench leaderboards. The column ``Origin'' indicates which leaderboard the model comes from. Note there are no RobustBench leaderboards for $l_2$ constrained adversarial attacks on CIFAR100 and ImageNet trained models. The best results for each column in each block are highlighted in bold.}
\label{tab-RobustBench_DAR}
      \setlength{\tabcolsep}{2.1pt}
     \resizebox{\ifdim\width>\linewidth\linewidth\else\width\fi}{!}{%
     \begin{tabular}{lcllC{15mm}C{15mm}C{15mm}C{15mm}C{15mm}C{15mm}}
        \toprule
      Origin     & Data-set &  Network      & Num.    & {Clean} &	{Corrupt}	& {Advers.} &	{Novel}	& {Unrecog.} & {Mean}\\
                 &            &  Architecture & Params. & DAR (\%) &DAR (\%) &DAR (\%) &DAR (\%) &DAR (\%) &{DAR (\%)}\\  \midrule
      our        &\multirow{14}{*}{\rotatebox{90}{CIFAR10}} &5-3-225   &11.1M  & 92.87&	86.36&\B90.41&\B78.09&\B    98.02&\B    89.15\\[2mm]

RobustBench&   &WideResNet18-2 &\multirow{2}{*}{268M}        & \multirow{2}{*}{\B94.16}&\multirow{2}{*}{86.63}&\multirow{2}{*}{40.76}&\multirow{2}{*}{71.32}&\multirow{2}{*}{68.59}&\multirow{2}{*}{72.29}\\
corruptions&   &\citet{Diffenderfer_etal21}&&&&&&& \\[1mm]
                                     
RobustBench&   &ResNet18 &\multirow{2}{*}{11.7M}               & \multirow{2}{*}{93.31}&\multirow{2}{*}{\B89.69}&\multirow{2}{*}{19.73}&\multirow{2}{*}{56.40}&\multirow{2}{*}{54.56}&\multirow{2}{*}{62.74}\\
corruptions&   &\citet{Kireev_etal22}&&&&&&&\\[1mm]
                                     
RobustBench&   &WideResNet70-16 &\multirow{2}{*}{267M}       & \multirow{2}{*}{93.53}&\multirow{2}{*}{88.70}&\multirow{2}{*}{82.80}&\multirow{2}{*}{73.49}&\multirow{2}{*}{48.53}&\multirow{2}{*}{77.41}\\
advers. ($l_2$)& &\citet{Wang_etal23}&&&&&&& \\[1mm]
                                     
RobustBench&   &WideResNet70-16 &\multirow{2}{*}{267M}       & \multirow{2}{*}{93.80}&\multirow{2}{*}{88.15}&\multirow{2}{*}{79.15}&\multirow{2}{*}{75.82}&\multirow{2}{*}{39.12}&\multirow{2}{*}{75.21}\\
advers. ($l_2$)&  &\citet{Rebuffi_etal21}&&&&&&&\\[1mm]
                                     
RobustBench&  &WideResNet94-16 &\multirow{2}{*}{366M}       &  \multirow{2}{*}{92.04}&\multirow{2}{*}{86.72}&\multirow{2}{*}{86.36}&\multirow{2}{*}{59.26}&\multirow{2}{*}{21.21}&\multirow{2}{*}{69.12}\\
advers. ($l_\infty$)& &\citet{Bartoldson_etal24}&&&&&&& \\[1mm]
                                     

RobustBench&  &WideResNet28-10 &\multirow{2}{*}{36.5M}       &  \multirow{2}{*}{91.55}&\multirow{2}{*}{86.16}&\multirow{2}{*}{83.20}&\multirow{2}{*}{59.89}&\multirow{2}{*}{21.43}&\multirow{2}{*}{68.45}\\
advers. ($l_\infty$)&  &\citet{Cui_etal23}   &&&&&&&\\[1mm]     

\midrule
our        &\multirow{14}{*}{\rotatebox{90}{CIFAR100}}&  5-3-210 &11.1M& 81.95&	77.20&\B70.16&	46.19&\B 77.56&\B	70.61\\[2mm]

RobustBench&  &WideResNet18-2 &\multirow{2}{*}{268M}       &\multirow{2}{*}{\B84.72}&\multirow{2}{*}{\B80.56}&\multirow{2}{*}{20.14}&\multirow{2}{*}{50.63}&\multirow{2}{*}{50.70}&\multirow{2}{*}{57.35}\\
corruptions & &\citet{Diffenderfer_etal21} &&&&&&&\\[1mm]

RobustBench&  &ResNet18 &\multirow{2}{*}{11.7M}               &\multirow{2}{*}{81.99}&\multirow{2}{*}{78.02}&\multirow{2}{*}{12.41}&\multirow{2}{*}{\B52.05}&\multirow{2}{*}{72.63}&\multirow{2}{*}{59.42}\\
corruptions & &\citet{Modas_etal22} &&&&&&&\\[1mm]     

RobustBench&  &ResNeXt29 32x4d &\multirow{2}{*}{6.9M}     & \multirow{2}{*}{83.71}&\multirow{2}{*}{74.80}&\multirow{2}{*}{9.04}&\multirow{2}{*}{39.68}&\multirow{2}{*}{28.56}&\multirow{2}{*}{47.16}\\
corruptions&  &\citet{Hendrycks_etal20} &&&&&&&\\[1mm]   

RobustBench&  &WideResNet70-16 &\multirow{2}{*}{267M}      &\multirow{2}{*}{80.38}&\multirow{2}{*}{72.74}&\multirow{2}{*}{63.99}&\multirow{2}{*}{42.86}&\multirow{2}{*}{5.94}&\multirow{2}{*}{53.18}\\
advers. ($l_\infty$)&  &\citet{Wang_etal23} &&&&&&&\\[1mm]      

RobustBench&  &WideResNet28-10 &\multirow{2}{*}{36M}       &  \multirow{2}{*}{80.12}&\multirow{2}{*}{73.62}&\multirow{2}{*}{62.32}&\multirow{2}{*}{45.58}&\multirow{2}{*}{7.96}&\multirow{2}{*}{53.92}\\
advers. ($l_\infty$)&  &\citet{Cui_etal23}   &&&&&&&\\[1mm]     

RobustBench&  &XCiT-L12 &\multirow{2}{*}{103M}             &  \multirow{2}{*}{75.95}&\multirow{2}{*}{72.25}&\multirow{2}{*}{58.85}&\multirow{2}{*}{34.18}&\multirow{2}{*}{60.11}&\multirow{2}{*}{60.27}\\
advers. ($l_\infty$)&  &\citet{Debenedetti_etal23}  &&&&&&&\\

\midrule
      our        &\multirow{10}{*}{\rotatebox{90}{ImageNet1k}}&   7-3-40-exp & 26.3M & 77.32&	74.57&\B	79.40&	53.73&\B	99.64&	76.93\\[2mm]
RobustBench&  &DeiT Base     &\multirow{2}{*}{86.6M}         &\multirow{2}{*}{\B85.75}&	\multirow{2}{*}{74.69}&	\multirow{2}{*}{32.10}&	\multirow{2}{*}{46.50}&	\multirow{2}{*}{99.61}& \multirow{2}{*}{67.73}\\
corruptions&  &\citet{Tian_etal22} &&&&&&&\\[1mm]

RobustBench&  &ResNet50      &\multirow{2}{*}{25.6M}         &\multirow{2}{*}{80.90}&	\multirow{2}{*}{74.77}&	\multirow{2}{*}{15.75}&	\multirow{2}{*}{37.83}&	\multirow{2}{*}{58.85}&	\multirow{2}{*}{53.62}\\
corruptions&  &\citet{Hendrycks_etal21} &&&&&&&\\[1mm]

RobustBench&  &Swin-L       &\multirow{2}{*}{197M}          &\multirow{2}{*}{83.34}&\multirow{2}{*}{78.92}&\multirow{2}{*}{72.40}&\multirow{2}{*}{\B55.18}&\multirow{2}{*}{97.52}&\multirow{2}{*}{\B77.47}\\
advers. ($l_\infty$) & &\citet{Liu_etal25}&&&&&&&\\[1mm]

RobustBench&  &ConvNeXt-L + ConvStem &\multirow{2}{*}{198M} &\multirow{2}{*}{79.80}&\multirow{2}{*}{\B79.26}&\multirow{2}{*}{76.10}&\multirow{2}{*}{46.70}&\multirow{2}{*}{98.98}&\multirow{2}{*}{76.17}\\
advers. ($l_\infty$)&  &\citet{Singh_etal23} &&&&&&&\\[1mm]
\hline
      \end{tabular}}
\end{center}
\end{table}

\subsection{Comparison to Leading RobustBench Models}
\label{sec-results_robustbench}

Modern DNNs are extremely sample-inefficient, and hence, require a large variety of training data and many training epochs. It might, therefore, be objected that the comparison performed in the preceding section still uses weak baselines that have not been trained to produce the best possible performance. To address this concern we compare the performance of our basic architecture composed of the proposed building-blocks to a range of SOTA models. These models were taken from the RobustBench \citep{Croce_etal21} leaderboards. 
A selection of the top performing models were chosen from each RobustBench leaderboard (the same models also appear at or near the top of another robustness benchmark \citep{Li_etal24}). 
The selected models included the highest ranked model for which pre-trained weights were available from the RobustBench repository\footnote{\url{https://github.com/RobustBench/robustbench}} on June 22$^{nd}$ 2026. Other tested models were selected from the top-10 of each leaderboard with a bias towards the highest ranked models, but lower ranked ones were sometimes selected to increase the diversity of network architectures and training methods used.

There are no restrictions on how  the RobustBench models are trained. Such models often use data additional to the official training-set, plus many different data augmentation techniques, carefully optimised training schedules and very long training times.
For the STEAL networks the only changes made were an increase in the number of channels, and the introduction of stronger data augmentations (\cref{sec-training_setup}). For the ImageNet1k-trained STEAL Net the number of training epochs was also increased to 200. With these basic changes the proposed models are competitive with the RobustBench models and consistently outperform them when tested with adversarial and unrecognisable data (\cref{tab-RobustBench_DAR}). Hence, even in comparison to the best machine learning models STEAL networks are far less prone to making confidently wrong predictions.

\subsection{Analysis of Confidence Scores}
\label{sec-results_confidence}

\begin{figure*}[tbp]
  \tabcolsep0pt
  \begin{tabular}{lrrrrr}
    &\multicolumn{1}{c}{clean}&\multicolumn{1}{c}{corrupt}&\multicolumn{1}{c}{adversarial}&\multicolumn{1}{c}{novel}&\multicolumn{1}{c}{unrecognisable}\\
        \rotatebox{90}{\hspace*{9mm}our}&
    \includegraphics[height=0.15\textwidth]{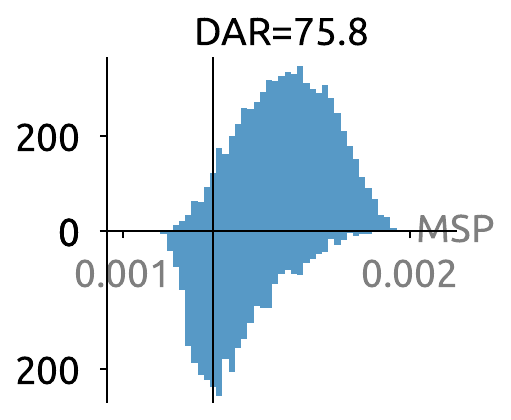}&
    \includegraphics[height=0.15\textwidth]{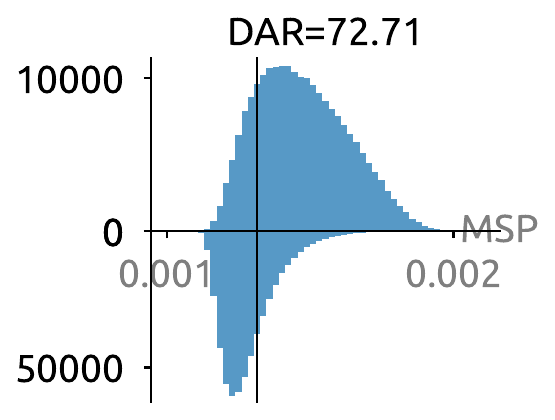}&
    \includegraphics[height=0.15\textwidth]{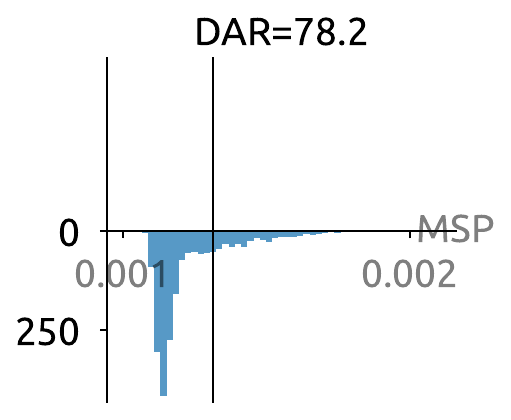}&
    \includegraphics[height=0.15\textwidth]{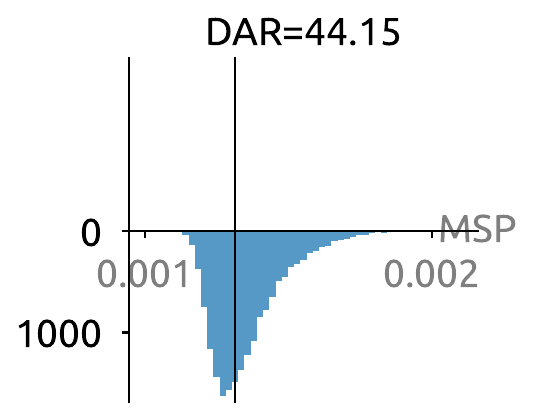}&
    \includegraphics[height=0.15\textwidth]{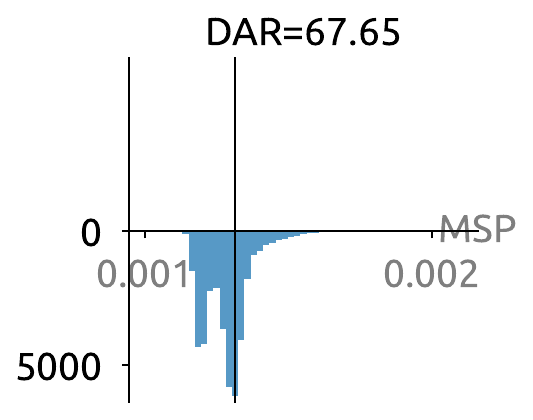}\\
        \rotatebox{90}{\hspace*{9mm}std}&
    \includegraphics[height=0.15\textwidth]{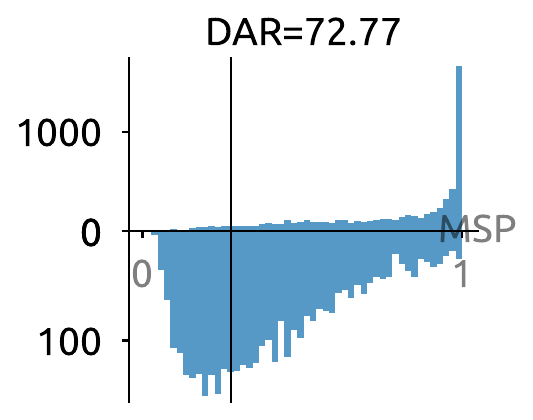}&
    \includegraphics[height=0.15\textwidth]{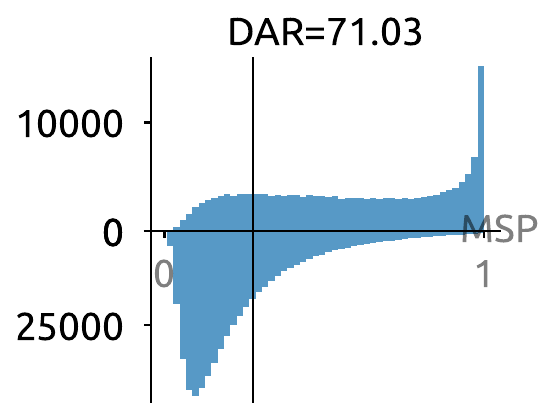}&
    \includegraphics[height=0.15\textwidth]{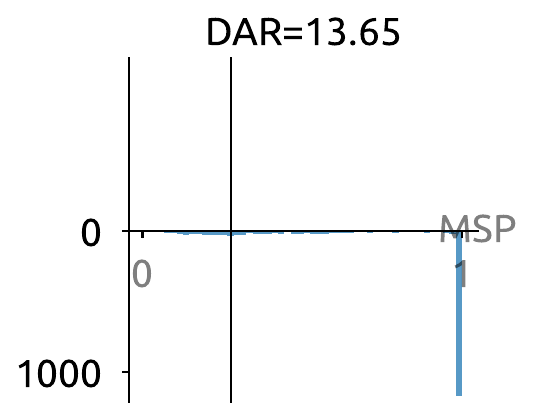}&
    \includegraphics[height=0.15\textwidth]{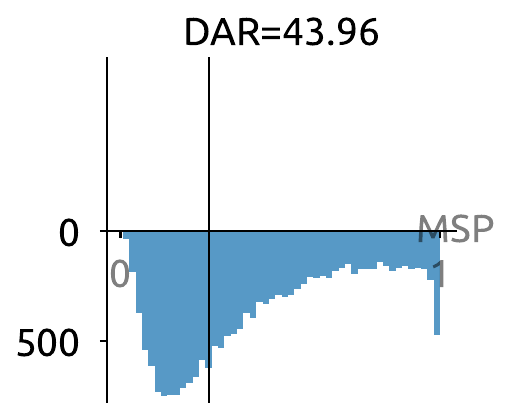}&
    \includegraphics[height=0.15\textwidth]{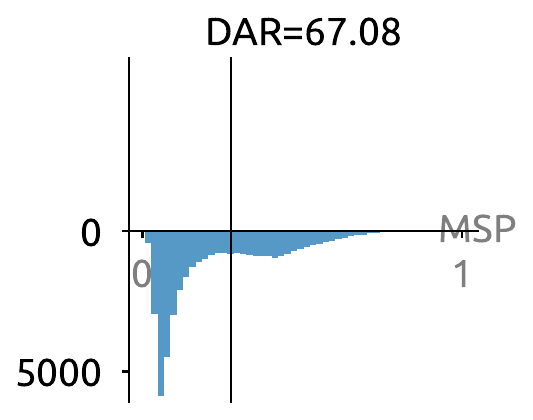}
    \end{tabular}
    \caption{Prediction confidence distributions for different types of test data. Produced for networks trained on ImageNet1k with basic data augmentations. The format of each sub-figure is described in the caption of \cref{fig-MSP_hist_truth_key}.}
    \label{fig-histograms_MSP_augbasic}
\end{figure*}

For the networks
trained on ImageNet1k and assessed in \cref{tab-results_DAR_all}, we plotted histograms showing the distribution of confidence scores produced by the samples from each type of test data (see \cref{fig-histograms_MSP_augbasic}).  While the equivalent models composed of the different building-blocks produce very different confidence distributions, there are similarities in the relative distribution of the clean, corrupt, unknown, and unrecognisable data produced by a single model. Specifically, these distributions all heavily overlap with each other for both STEAL Nets and equivalent CNNs.

In contrast, the confidence scores produced by adversarial data are very different for the two architectures. For the standard CNN, adversarial samples are classified (incorrectly) with very high confidence (on average higher confidence that is produced for the clean test data). In contrast, for the proposed architecture, adversarial samples produce confidence scores that are lower on average than those for the clean data. As a result, for the proposed architecture, the distribution of confidence scores for the adversarial data overlaps with the confidence distributions for the other test data-types. Hence, for the standard CNN adversarial samples are a unique type of data with properties unlike any other type of test data. In contrast, for the STEAL network, adversarial samples have properties similar to those of other data types, especially data that has a different distribution compared to the training data.

\subsection{Analysis of Adversarial Robustness}
\label{sec-results_adversarial}

The networks
assessed above have not been trained to be adversarially robust, and as a result typically have a robust accuracy near zero (\cref{tab-results_conventional_all,tab-results_conventional_ImageNet_basic,tab-RobustBench_conventional}). The non-zero performance recorded with DAR for adversarial attacks on STEAL Nets is, therefore, primarily due to the rejection of adversarially-perturbed samples. The analysis in the previous section explains why STEAL networks composed of the proposed building blocks can successfully reject adversarial samples (because this data produces predictions with low confidence). In contrast, networks composed of the standard building blocks fail to reject adversarial samples (because this data produces predictions with high confidence).

\begin{figure*}[tb]
  \begin{center}
  \tabcolsep0pt
  \begin{tabular}{lrrrr}
    &\multicolumn{1}{c}{$\epsilon=0.001$}&\multicolumn{1}{c}{$\epsilon=0.003$}&\multicolumn{1}{c}{$\epsilon=0.005$}\\
        \rotatebox{90}{\hspace*{9mm}our}&
    \includegraphics[height=0.15\textwidth]{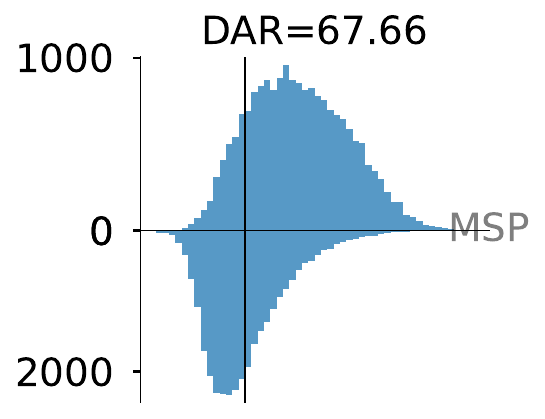}&
    \includegraphics[height=0.15\textwidth]{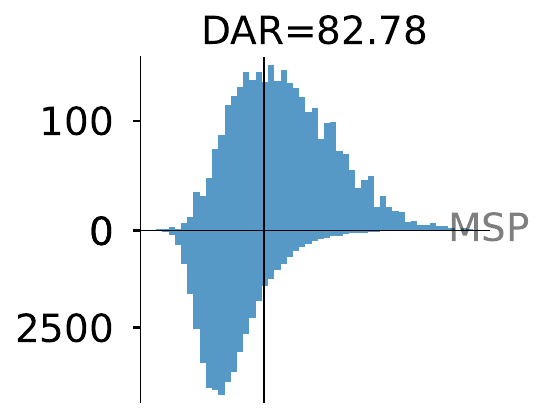}&
    \includegraphics[height=0.15\textwidth]{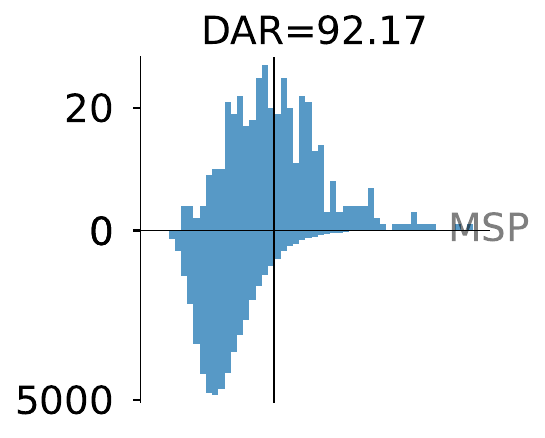}&
    \hspace*{6mm}\includegraphics[height=0.16\textwidth]{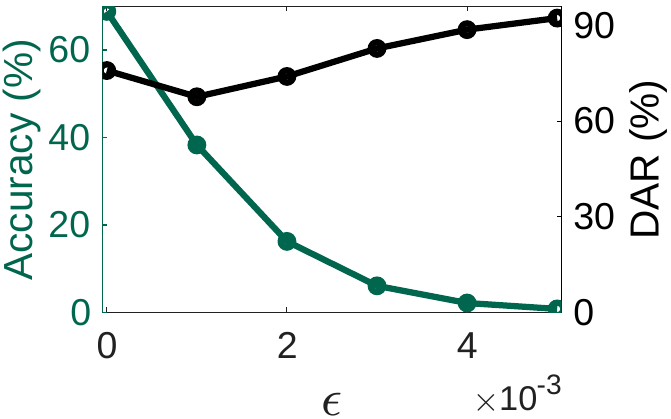}\\
         \rotatebox{90}{\hspace*{9mm}std}&
   \includegraphics[height=0.15\textwidth]{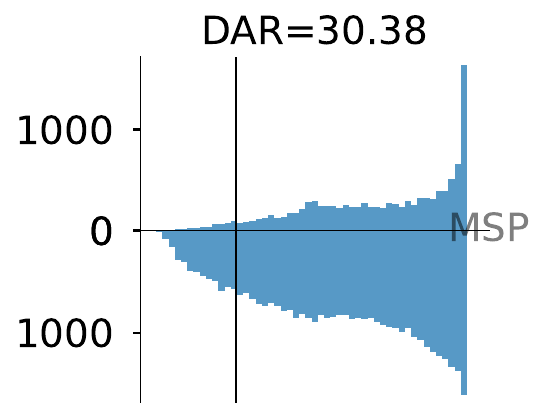}&
    \includegraphics[height=0.15\textwidth]{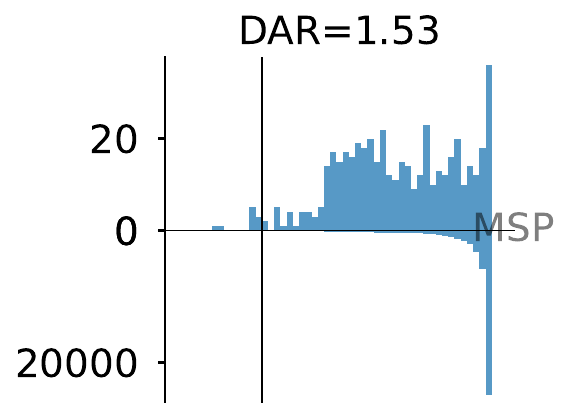}&
    \includegraphics[height=0.15\textwidth]{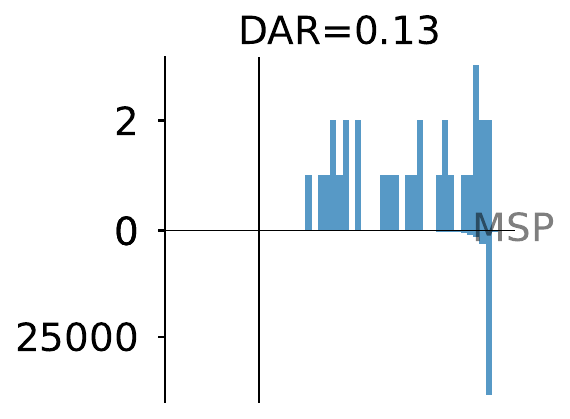}&
   \includegraphics[height=0.16\textwidth]{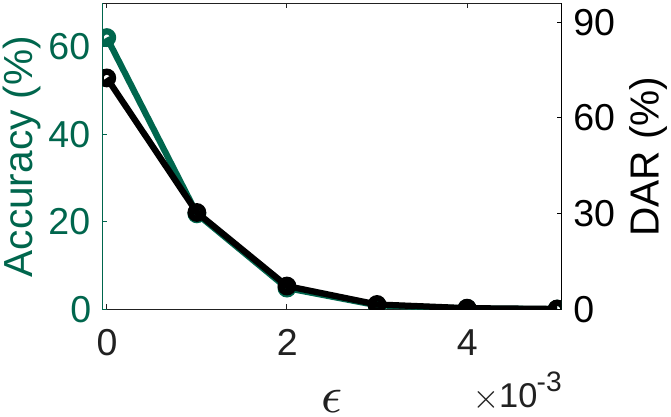}
    \end{tabular}
    \caption{Prediction confidence distributions for PGD$_{l_\infty}^{100}$ adversarial attacks of different strengths. Produced for networks trained on ImageNet1k with basic data augmentations. The format of each sub-figure is described in the caption of \cref{fig-MSP_hist_truth_key}. The plots in the right-hand column show the classification accuracy and DAR for adversarial attacks of different strengths. }
    \label{fig-adversarial_confidence_MSP_PGDLinf}
  \end{center}
\end{figure*}

The shift in confidence can be observed by plotting the confidence distributions for different attack strengths (\cref{fig-adversarial_confidence_MSP_PGDLinf}). For weak attacks both network architectures produce confidence distributions that strongly overlap that of the clean data, and these samples with small adversarial perturbations are often classified correctly (as shown by the accuracy at low $\epsilon$ in the right-hand column of  \cref{fig-adversarial_confidence_MSP_PGDLinf}. As attack strength increases, for the STEAL net the accuracy falls, but the prediction confidence also reduces, meaning that these samples can be rejected and the DAR remains high. In contrast, for the standard CNN, increasing the attack strength increases the confidence score, meaning that both accuracy and DAR decrease. 
The above experiments were performed using 100 steps of PGD with the maximum allowed perturbation constrained by the $l_\infty$-norm (denoted as PGD$_{l_\infty}^{100}$ for brevity) to be less than the perturbation budget ($\epsilon$). Very similar results (not shown) were observed when using PGD$_{l_2}^{100}$.

\begin{figure*}[tbp]
  \tabcolsep0pt
  \begin{tabular}{lrrrrr}
    &\multicolumn{1}{c}{clean}&\multicolumn{1}{c}{corrupt}&\multicolumn{1}{c}{adversarial}&\multicolumn{1}{c}{novel}&\multicolumn{1}{c}{unrecognisable}\\
     \multicolumn{4}{l}{\small \bf Strong Augmentations + weak AT}\\
        \rotatebox{90}{\hspace*{9mm}our}&
    \includegraphics[height=0.15\textwidth]{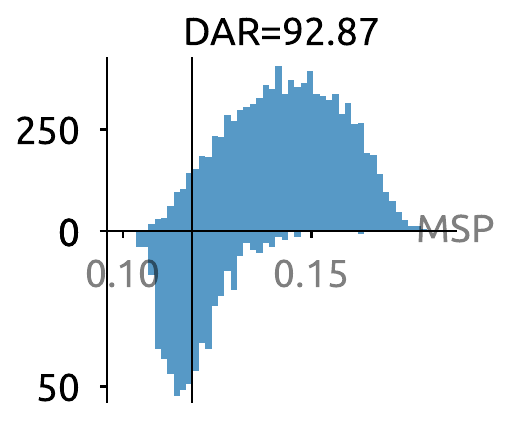}&
    \includegraphics[height=0.15\textwidth]{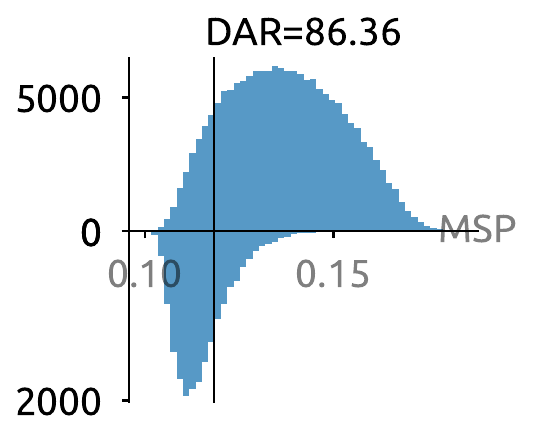}&
    \includegraphics[height=0.15\textwidth]{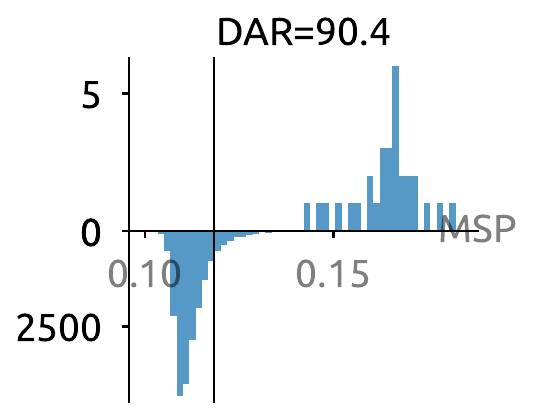}&
    \includegraphics[height=0.15\textwidth]{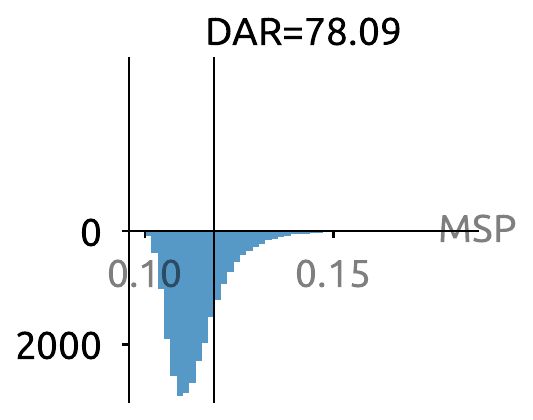}&
    \includegraphics[height=0.15\textwidth]{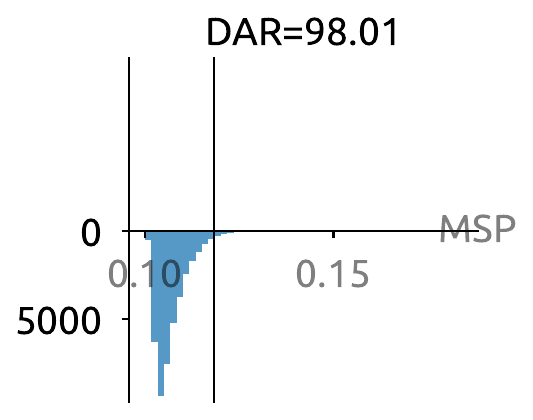}\\
        \rotatebox{90}{\hspace*{9mm}std}&
        \hspace*{0.4mm}
    \includegraphics[height=0.15\textwidth]{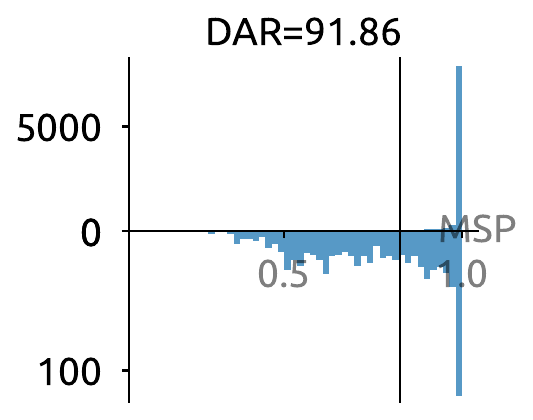}&
    \includegraphics[height=0.15\textwidth]{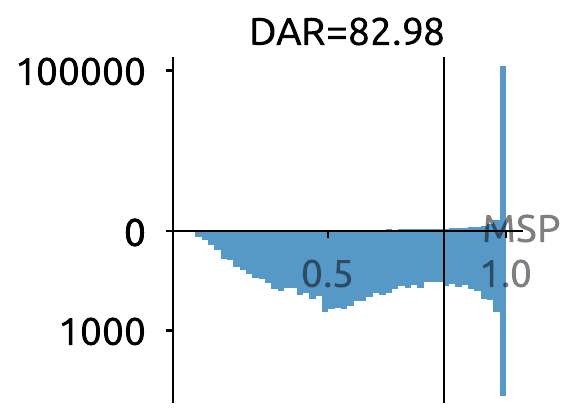}&
    \includegraphics[height=0.15\textwidth]{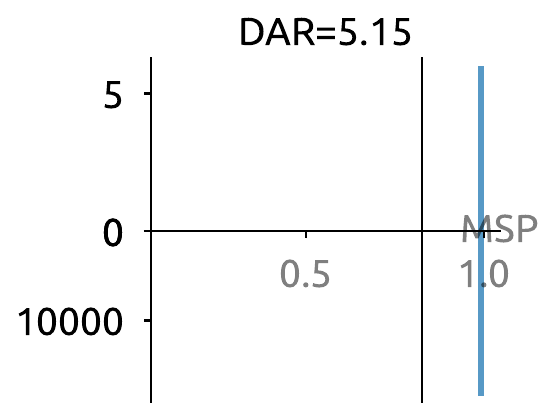}&
    \includegraphics[height=0.15\textwidth]{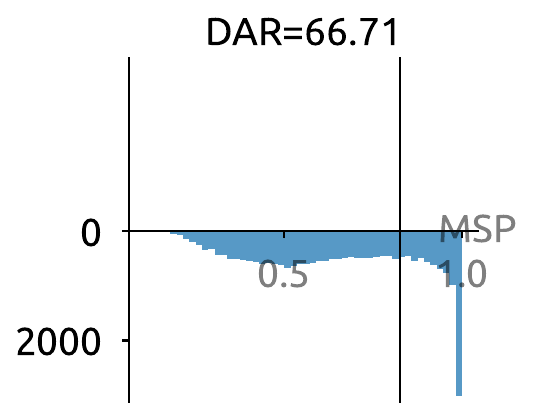}&
    \includegraphics[height=0.15\textwidth]{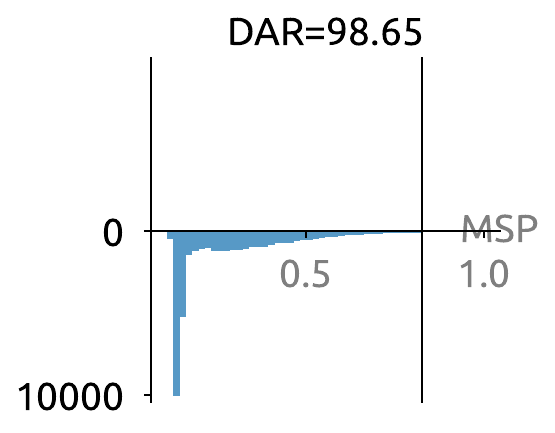}\\
     \multicolumn{4}{l}{\small \bf Strong Augmentations + strong AT}\\
        \rotatebox{90}{\hspace*{9mm}our}&
    \includegraphics[height=0.15\textwidth]{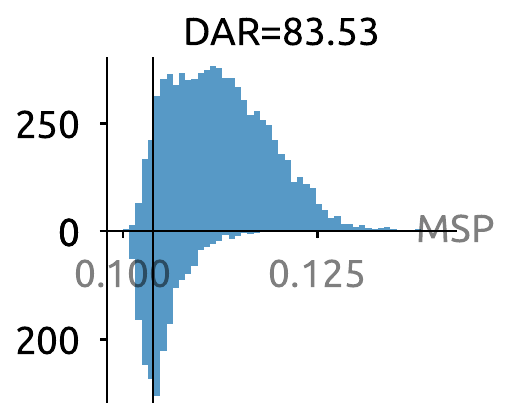}&
    \includegraphics[height=0.15\textwidth]{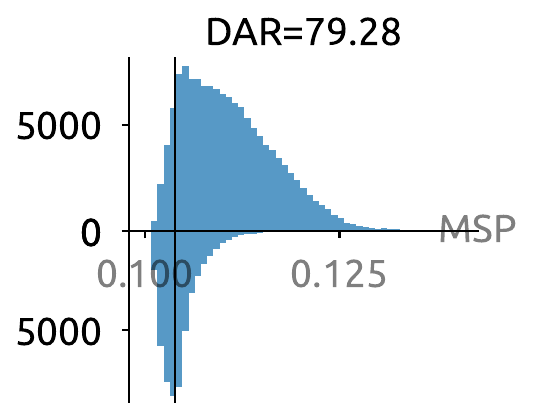}&
    \includegraphics[height=0.15\textwidth]{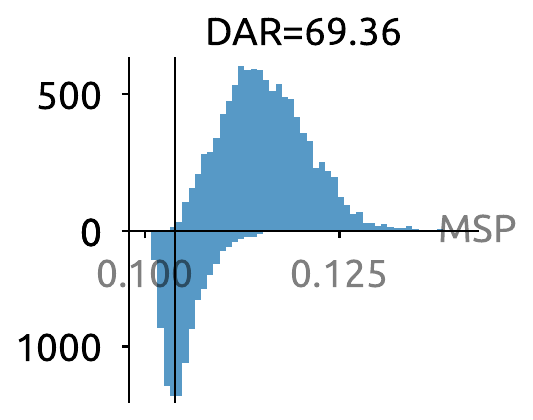}&
    \includegraphics[height=0.15\textwidth]{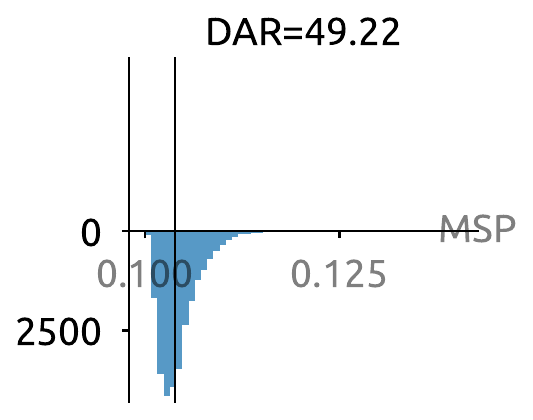}&
    \includegraphics[height=0.15\textwidth]{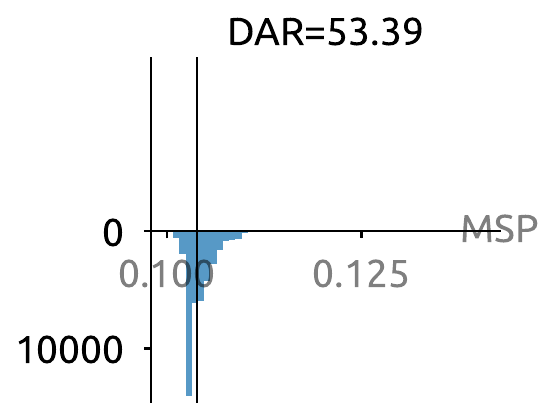}\\
        \rotatebox{90}{\hspace*{9mm}std}&
        \hspace*{0.4mm}
    \includegraphics[height=0.15\textwidth]{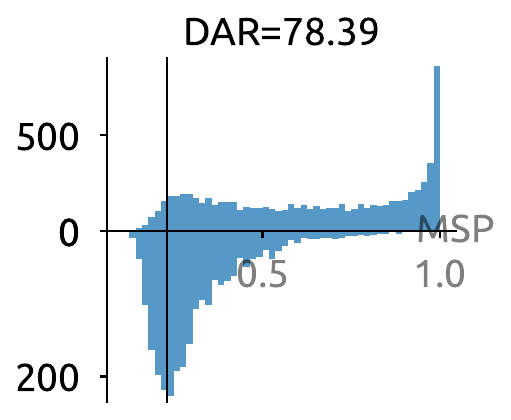}&
    \includegraphics[height=0.15\textwidth]{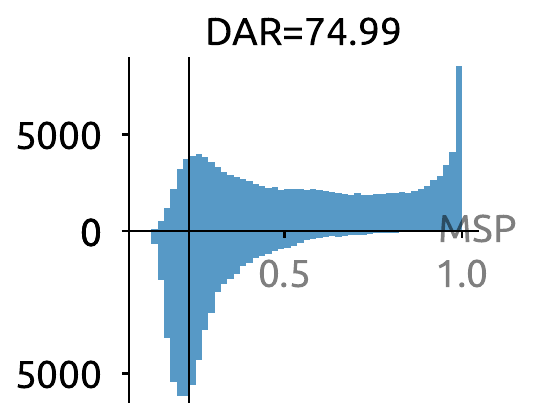}&
    \includegraphics[height=0.15\textwidth]{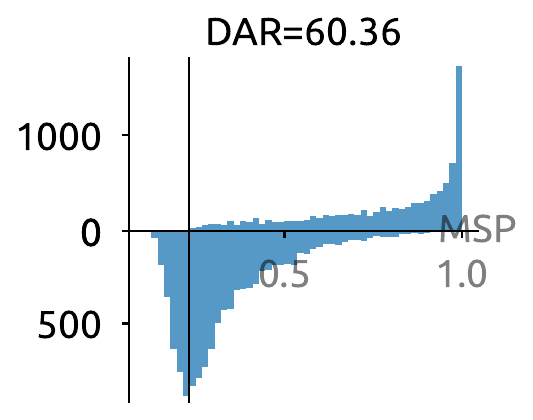}&
    \includegraphics[height=0.15\textwidth]{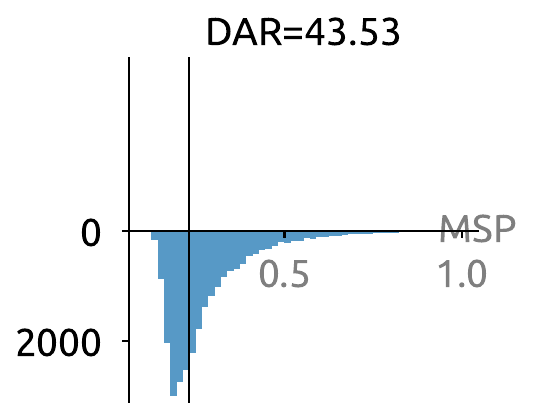}&
    \includegraphics[height=0.15\textwidth]{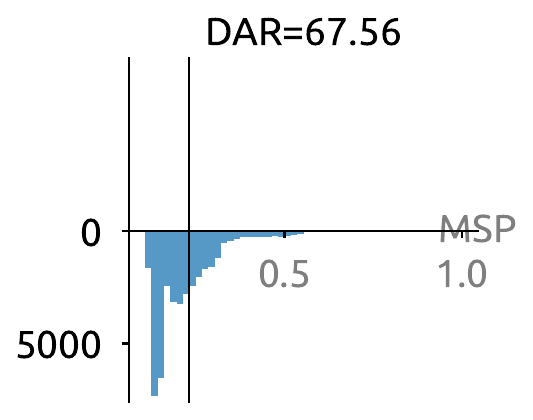}\\
    \end{tabular}
    \caption{Prediction confidence distributions for different types of test data. Produced for networks trained on CIFAR10 with strong data augmentations and adversarial training (AT). The format of each sub-figure is described in the caption of \cref{fig-MSP_hist_truth_key}.}
    \label{fig-histograms_MSP_AT}
\end{figure*}

\subsection{Effects of Adversarial Training}
\label{sec-results_AT}

  Human classification decisions are unaffected by adversarial attacks of the kind we have used. In contrast, STEAL nets are sensitive to them. More specifically, our networks are poor at generalising to adversarial samples. A standard machine learning technique to improve generalisation is to augment the training data with exemplars of the kind that better performance is desired for. A standard technique to improve the ability of standard DNNs to correctly classify adversarial samples is Adversarial Training (AT).   
  AT modifies training samples using perturbations generated by an adversarial attack.
  In this section we explore if AT can help STEAL Nets to correctly classify adversarial samples.
  
  The strong training augmentation scheme used with the proposed network architecture to produce the results shown in \cref{tab-RobustBench_DAR} applies FGSM \citep{Goodfellow_etal15}, a single-step attack method, to 20\% of the training images to produce training-time adversarial images. However, this very weak method of AT does not increase adversarial robustness noticeably above zero (\cref{tab-RobustBench_conventional}). This is expected as it well-known that AT implemented with single-step attacks leaves models vulnerable to stronger attacks which use an iterative method to generate adversarial samples \citep{Kurakin_etal17b}, as are used in AutoAttack. 
The histograms of confidence scores produced by networks trained on CIFAR10 using this method (\cref{fig-histograms_MSP_AT}, top) have strong similarity to those produced by networks trained on ImageNet1k using basic training augmentations (\cref{fig-histograms_MSP_augbasic}). 

Performing adversarial training using multi-step attacks, such as Projected Gradient Descent (PGD), to generate the training-time adversarial images produces much more robust classifiers \citep{Madry_etal18}. As a result PGD-AT has become a standard benchmark against which all other methods are judged. We tested the effects of this strong form of AT by replacing FGSM applied to 20\% of the images with PGD$_{l_\infty}^{7}$ ($\epsilon=\frac{8}{255}$) applied to 100\% of the training images. For a network composed of the proposed building-blocks the main effect of strong AT was to reduce the average confidence score produced by the clean and corrupt test-data, while increasing the score for (AutoAttack) adversarial test-data (\cref{fig-histograms_MSP_AT}, third row). The increased overlap in confidence scores for all types of test data results in a wholesale reduction in DAR.
For a network composed of standard building-blocks, as for the proposed architecture, strong AT reduces the average confidence score produced by the clean and corrupt test-data. However, unlike for the proposed architecture, strong AT also reduces the confidence produced by adversarial and unknown samples (\cref{fig-histograms_MSP_AT}, last row). In terms of quantitative performance, the effect of strong AT is to reduce DAR for all data-types except adversarial test-data. Despite this rise in adversarial DAR for the standard network it is still less adversarially robust than the proposed network architecture.

In terms of conventional evaluation metrics, there is a well-known trade-off between clean accuracy and adversarial accuracy \citep{Zhang_etal19,Tsipras_etal19,Madry_etal18,Kannan_etal18}. For the proposed networks this trade-off is less severe: a gain in adversarial accuracy of 51.51\% is accompanied by a reduction of clean accuracy of 13.9\% (\cref{tab-results_conventional_CIFAR10_AT}). In comparison, for a size-matched standard network the gain in adversarial robustness is lower (46.38\%) while the fall in clean accuracy is greater (17.02\%). Overall, these result, although preliminary, suggest that it should be easier to train adversarial robust networks built from the proposed components than standard ones.


An adversarial attack performed on one DNN is typically also an effective attack on a separate DNN \citep{Szegedy_etal14}. We were interested to know if such attack transferability would also occur between the proposed and standard architectures. For both architectures transfer attacks from one model to another of the same type are slightly less effective than when the attack is generated for the same model as is attacked (\cref{tab-results_CIFAR10_AT_transferability}). In contrast, when adversarial attacks are transferred from one type of architecture to another there is some asymmetry in the results. Specifically, transferring adversarial samples from the proposed architecture to the standard one is much more effective than the reverse. In other words, a STEAL Net is more robust to adversarial attacks produced using a CNN than the CNN is to adversarial attacks transferred from the STEAL Net.

\begin{table*}[tbp]\centering
  \caption{Transferablility of adversarial attacks. Experiments performed using networks trained on the CIFAR10 data-set with Strong Augmentations + strong AT. Reported values are Adversarial Accuracy (\%).}   
    \label{tab-results_CIFAR10_AT_transferability}
    \tabcolsep5pt
    \resizebox{\ifdim\width>\linewidth\linewidth\else\width\fi}{!}{%
      \begin{tabular}{r cc}\toprule
Network used to create & \multicolumn{2}{c}{Attacked Network}\\
adversarial samples                   & our (v1) & std (v1)\\\midrule
same as attacked network               & 51.66    & 46.41  \\
our (v2)                               & 57.25    & 61.55        \\
std (v2)                               & 70.06    & 55.67        \\
\bottomrule\end{tabular}}\end{table*}

\begin{figure*}[tbp]
  \begin{center}
    \begin{tabular}{ccccc}
        &{our}&{std}&{our}&{std}\\
    \rotatebox{90}{\hspace*{13mm}\small MNIST}&
     \includegraphics[scale=0.37]{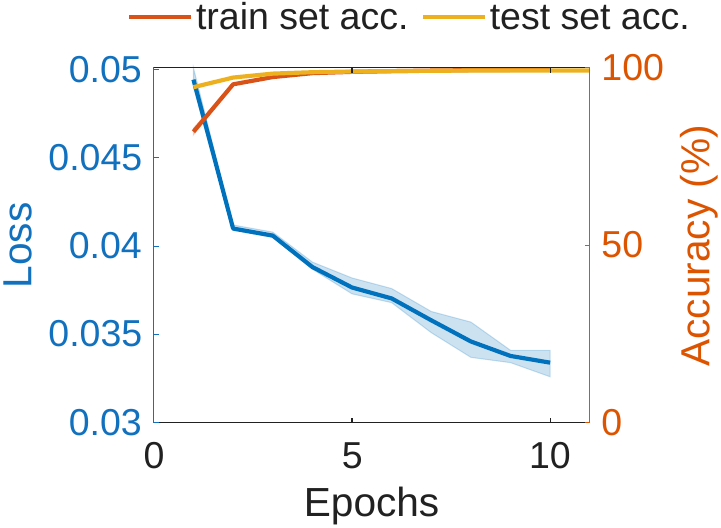}&
   \includegraphics[scale=0.37,trim=0 0 0 16, clip]{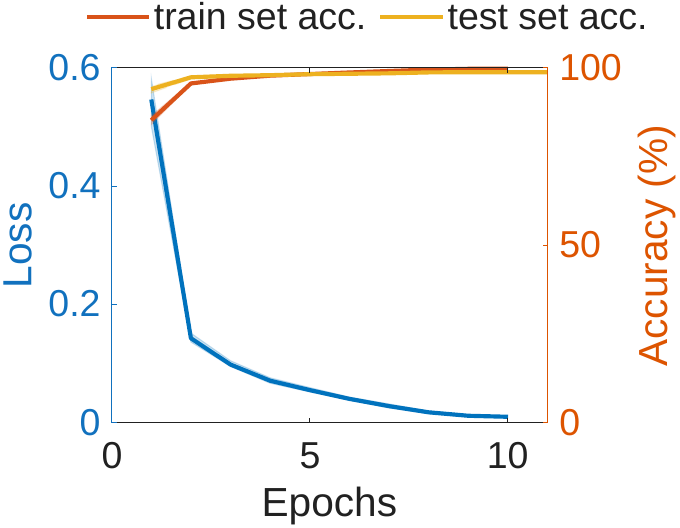}&
    \includegraphics[scale=0.25]{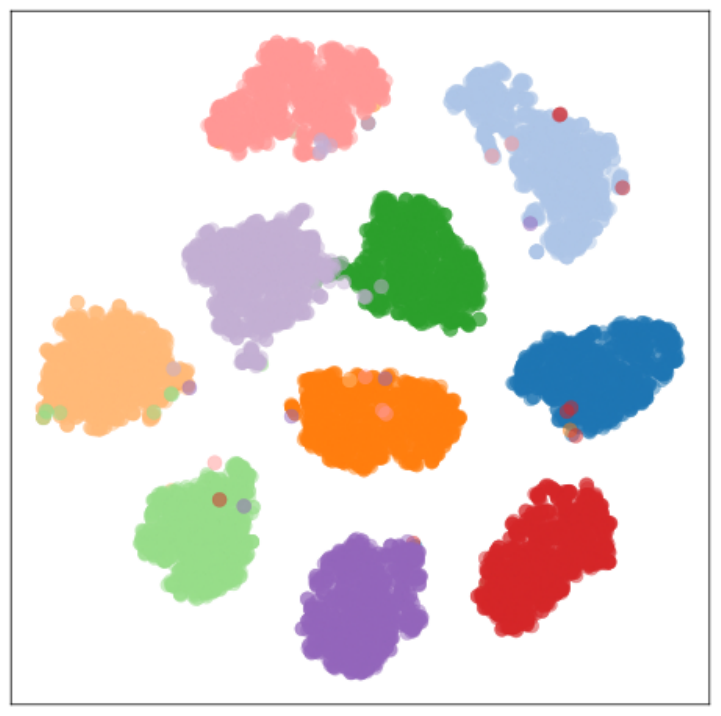}&
    \includegraphics[scale=0.25]{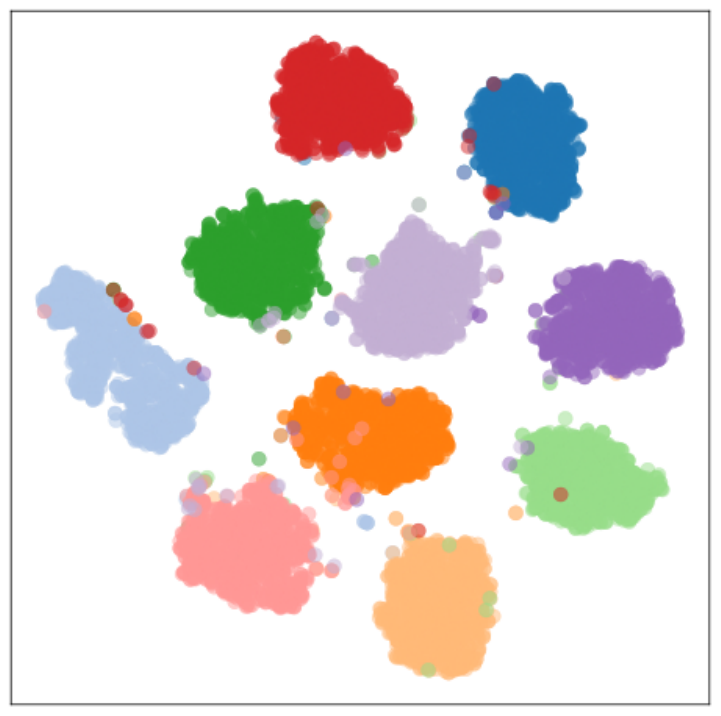}\\
    
    \rotatebox{90}{\hspace*{14mm}\small SVHN}&
    \includegraphics[scale=0.37,trim=0 0 0 16, clip]{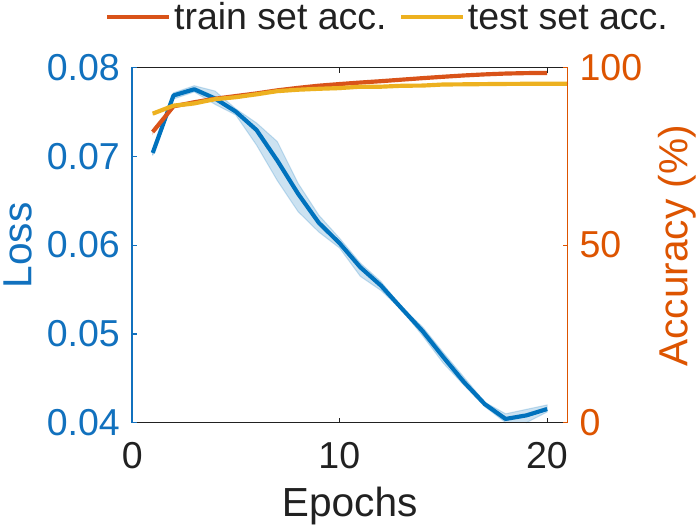}&
    \includegraphics[scale=0.37,trim=0 0 0 16, clip]{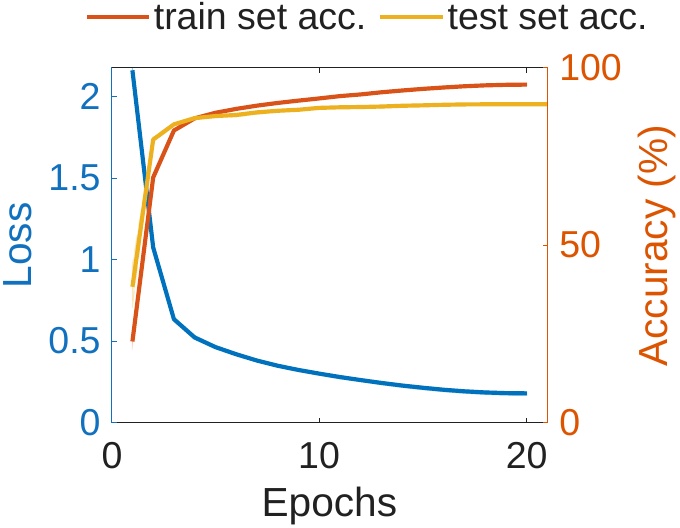}&
    \includegraphics[scale=0.25]{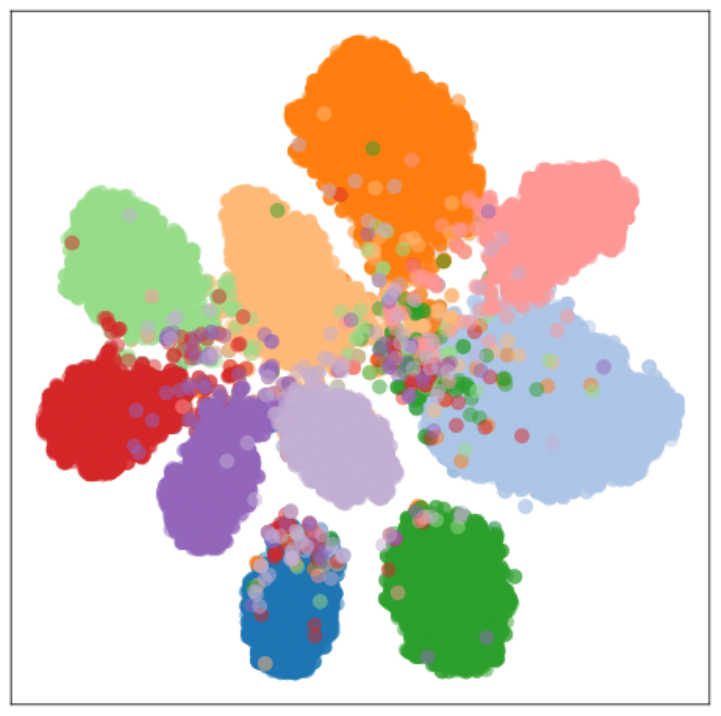}&
    \includegraphics[scale=0.25]{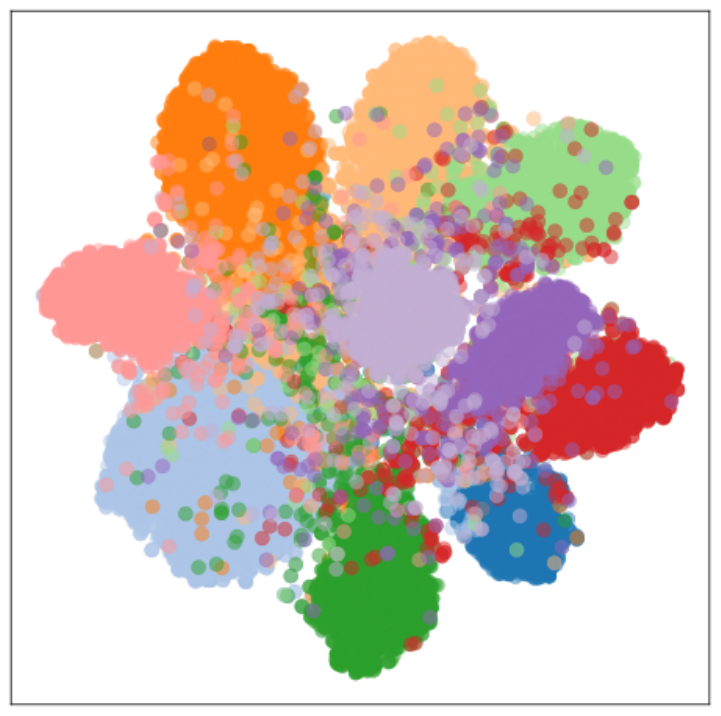}\\
    
    \rotatebox{90}{\hspace*{12mm}\small CIFAR10}&
    \includegraphics[scale=0.37,trim=0 0 0 16, clip]{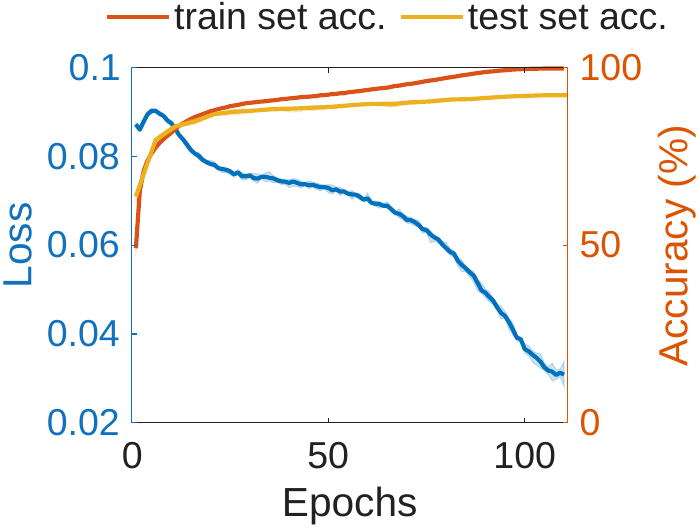}&
    \includegraphics[scale=0.37,trim=0 0 0 16, clip]{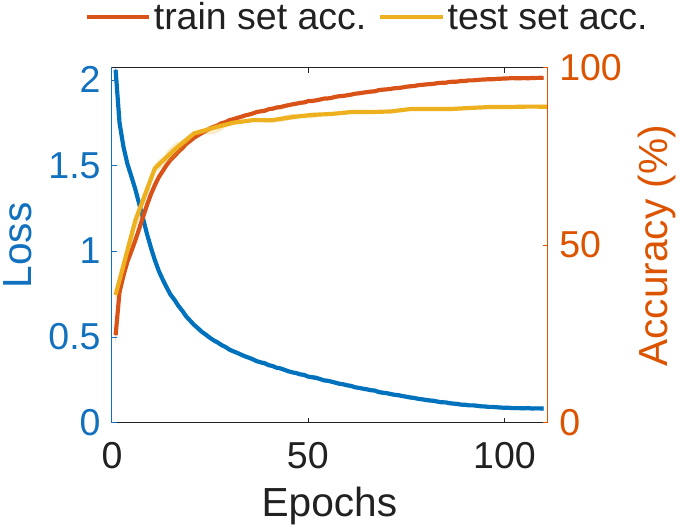}&
    \includegraphics[scale=0.25]{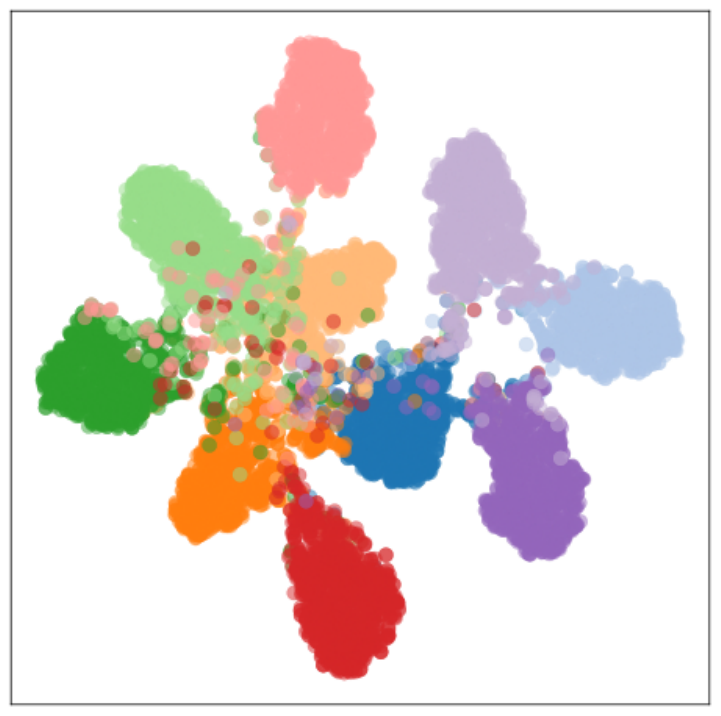}&
    \includegraphics[scale=0.25]{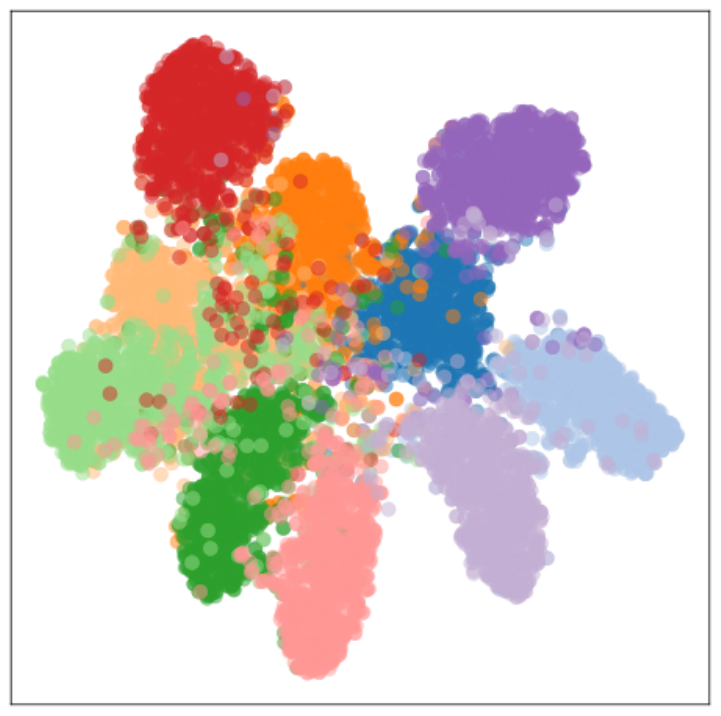}\\
    \end{tabular}
        \caption{Training dynamics (two left-hand columns) and final learnt t-SNE embeddings (two right-hand columns) for selected data-sets. Each graph in the two left-hand columns shows the change during training of the loss, the percentage clean accuracy on the training-set and on the test-set. The loss is measured against the left-hand vertical axis and the accuracy against the right-hand axis. The solid lines show the mean values over five trials, and the shaded regions (which are too narrow to be clearly visible) indicate the minimum and maximum values recorded in any of the five trials. Each graph in the two right-hand columns shows the location of test samples in the feature-space learnt by a network, with the colour of the dot indicating the class of the sample.}
    \label{fig-training_dynamics}
  \end{center}
\end{figure*}

\begin{figure}[tbp]
\begin{center}
\includegraphics[width=0.3\textwidth]{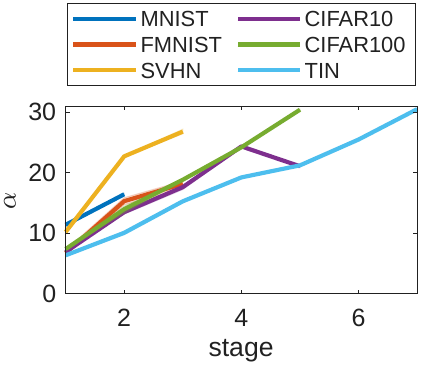}
\caption{The value of $\alpha$ learnt,  at the end of training, by the SCAM layer in each stage of a network. Results are for the same network architectures listed in \cref{tab-results_DAR_all}. The solid lines show the mean values over five trials, and shaded regions which are too narrow to be visible indicate the minimum and maximum values recorded in any of the five trials.}
\label{fig-alpha}
\end{center}
\end{figure}

\subsection{Analysis of Learning}
\label{sec-results_learning}

To check that both the standard and modified NN architectures used to produce the results in \cref{tab-results_DAR_all} were trained effectively, \cref{fig-training_dynamics} plots the values of various metrics over the course of training on three data-sets.
A major difference between CE and HEM loss is that the latter does not monotonically reduce over the whole course of training. This is to be expected, as only errors greater than the average contribute to the loss. Hence, it is possible that parameter updates during learning cause errors to move from just above the average to below the average. This will increase the mean of the remaining errors, and hence, the loss. This selection of the largest errors also causes HEM loss to be more variable.
It can be seen that loss decreases and accuracy (for both the training data and the test data) increases over the course of training for both network architectures, reaching a plateaux towards the end of training. In all cases, final train-set accuracy is higher than the final test-set accuracy indicating some degree of over-fitting. However, the over-fitting in the proposed models is no more severe than that in the standard models, despite weight decay being ineffective for ZINC and the lack of any other regularisation method (see \cref{sec-training_setup}).

The final feature-embeddings learnt by the different models, plotted using t-SNE \citep{HintonRoweis02}, show that the regions occupied by each class are better separated by STEAL networks
(\cref{fig-training_dynamics}, third column) compared to equivalent networks composed of standard building-blocks (\cref{fig-training_dynamics}, fourth column).

In STEAL Nets, as well as learning the synaptic weights in the ZINC layers, back-propagation is also used to learn the $\alpha$ values in the SCAM layers (see \cref{sec-SCAM}). The final values of $\alpha$ learnt across different repeats of training with the same data-set and network architecture are remarkably consistent (\cref{fig-alpha}). Furthermore, the general trend of $\alpha$ being smaller in the early layers, and larger in the later layers, is also consistent across training data-set and different network depths.

\begin{table}[tbp]
  \centering
  \tabcolsep6pt
  \caption{An evaluation of the effects of training set-up on clean accuracy (with the standard CIFAR10 test-set).
    `our' denotes STEAL networks composed of the proposed building-blocks (ZINC, SCAM, and HEM).
    `std' denotes size-matched CNNs composed of standard building-blocks (CC, Max pooling, and CE).
    The schedule `None' indicates that the learning rate was constant throughout training. For each training regime the best results across the two types of network is denoted in bold. For each type of network, the best results across all training regimes is indicated by underlining. Note: only one trial was performed in each condition, however, the results reported in \cref{tab-results_conventional_all} suggests that results should be fairly consistent across repeats.}
    \label{tab-training_setup}
      \begin{tabular}{cccccccccc}\toprule
 &                  & \multicolumn{8}{c}{Schedule} \\\cmidrule(lr){3-10}     
\multirow{2}{*}{\shortstack{Learning\\Rate}} & Optimiser\hspace*{2mm} &   \multicolumn{2}{c}{None} &\hspace*{1mm}& \multicolumn{2}{c}{Multi-Step} &\hspace*{1mm}& \multicolumn{2}{c}{One-Cycle}\\\cmidrule(lr){3-4}\cmidrule(lr){6-7}\cmidrule(lr){9-10}
           &         &our&std      &&our&std    &&our&std \\
\midrule
0.001 &SGD       &  68.51&\B80.70  &&  66.50 &\B80.93  &&  55.94&\B73.62 \\
      &RMSprop   &\B89.61&  85.53  &&\B90.60 &  86.99  &&\B91.19&  82.76 \\
      &AdamW     &\B90.54&  85.00  &&\B91.02 &  87.08  &&\B91.19&\underline{88.57} \\
      &Lion      &\B90.65&  10.00  &&\B91.26 &  10.00  &&\B90.90&  10.00 \\
      &Adan      &\B89.20&  85.46  &&\B90.42 &  86.75  &&\B89.97&  87.89\\
       \midrule                 
0.01  &SGD       &\B87.13&  86.96  &&  87.79 &\B88.37  &&  83.88 &\B87.32 \\
      &RMSprop   &  10.00&  10.00  &&  10.00 &  10.00  &&  10.00&  10.00 \\
      &AdamW     &\B89.33&  67.93  &&\B91.95 &  62.05  &&\B91.58&  84.96 \\
      &Lion      &\B89.31&  10.00  &&\B88.83 &  10.00  &&\B91.39&  10.00 \\
      &Adan      &\B90.35&  77.37  &&\B90.94 &  77.37  &&\B91.59&  86.29 \\
\midrule                 
0.1   &SGD       &\B89.45&  10.00  &&\B90.75 &  75.12  &&\B90.16 &  87.29 \\
      &RMSprop   &  10.00&  10.00  &&  10.00 &  10.00  &&  10.00&  10.00 \\
      &AdamW     &\B85.58&  10.00  &&\B91.79 &  10.00  &&\underline{\B92.59} &  10.00 \\
      &Lion      &  10.00&  10.00  &&  10.00 &  10.00  &&\B90.52&  10.00 \\
      &Adan      &  10.00&  10.00  &&  10.00 &  10.00  &&  10.00&  10.00 \\
\bottomrule
      \end{tabular}
\end{table}

\begin{table*}[tbp]\centering
  \caption{Evaluation of sensitivity to hyper-parameter values performed using the CIFAR10 data-set. The first row shows results obtained with the default hyper-parameter values and the last row shows results for a size-matched CNN built using standard components (these results also appear in \cref{tab-results_DAR_all} and are repeated here for convenience). The other rows show results when one hyper-parameter was changed from its default value while all other hyper-parameter remained unchanged. Bold text indicates the best performance on each metric.}
    \label{tab-results_DAR_sensitivity}
    \tabcolsep2.5pt
    \resizebox{\ifdim\width>\linewidth\linewidth\else\width\fi}{!}{%
      \begin{tabular}{lccccll ccccc *{1}{S[table-format=2.2(2)]}}\toprule
&\multicolumn{2}{c}{ZINC} &SCAM&HEM& Network        &Num.    &  Clean &Corrupt& Advers. &	Novel	& Unrecog. & {Mean $\pm$ Std}\\
&$\lambda$&$\gamma$&$\kappa$&$\mu$ & Arch.   &Params. &  DAR (\%) &DAR (\%) &DAR (\%) &DAR (\%) &DAR (\%) &{DAR (\%)}\\    \midrule
our&2 &  0.1  &8  & 0.1           &  5-3-44  &459k      &\B91.50&  80.26&  69.85&  66.32&\B92.25&\B80.04	(1.17)\\[1mm]
   &1 &  0.1  &8  & 0.1           &  5-3-44  &459k      &  91.34&  80.23&  66.64&  65.07&  83.14&  77.29	(1.48)\\
   &4 &  0.1  &8  & 0.1           &  5-3-44  &459k      &  91.43&	80.66&	67.92&	68.39&	89.03&	79.49	(1.48)\\[1mm]
   &2 &  0.05 &8  & 0.1           &  5-3-44  &459k      &  91.43&\B80.67&	66.34&	68.94&	89.41&	79.36	(1.12)\\
   &2 &  0.2  &8  & 0.1           &  5-3-44  &459k      &  91.45&	79.99&	70.32&	63.79&	73.19&	75.75	(3.07)\\[1mm]
   &2 &  0.1  &4  & 0.1           &  5-3-63  &458k      &  91.47&	80.34&	46.37&\B71.09&	88.83&	75.62   (1.02)\\
   &2 &  0.1  &16 & 0.1           &  5-3-30  &445k      &  91.22&	80.40&	46.65&	61.04&	83.04&	72.47	(1.43)\\[1mm]
   &2 &  0.1  &8  & 0.05          &  5-3-44  &459k      &  91.06&	80.27&	61.85&	63.44&	89.75&	77.27	(1.15)\\
   &2 &  0.1  &8  & 0.2           &  5-3-44  &459k      &  91.14&	79.95&\B72.39&  70.25&	85.52&	79.85	(1.09)\\[1mm]
std&\multicolumn{2}{c}{CC}&Max&CE &  5-3-128 &458k      &  89.49&  77.47&	5.01 &  47.25&	34.59&	50.76	(1.57)\\
\bottomrule\end{tabular}}\end{table*}

\begin{table*}[tbp]\centering
  \caption{Results for the ablation study  performed using the CIFAR10 data-set.  The top panel shows the effects of removing one of the proposed components (ZINC, SCAM, and HEM) and replacing it with the corresponding standard component (CC, Max pooling, and CE). The lower panel shows the effect of replacing components in a standard CNN with the proposed methods. $\dagger$\xspace In this condition, the parameter $\alpha$ used by SCAM was fixed to a value of one, as training failed when attempting to learn this parameter or when it was set to a higher value.}
    \label{tab-results_DAR_ablation}
    \tabcolsep2pt
    \resizebox{\ifdim\width>\linewidth\linewidth\else\width\fi}{!}{%
      \begin{tabular}{lllllll ccccc *{1}{S[table-format=2.2(2)]}}\toprule
&Integrat. &Pooling   &Loss &  Network        &Num.    & Training   & Clean &	Corrupt	& Advers. &	Novel	& Unrecog. & {Mean $\pm$ Std}\\
&Func.       &Func.     &Func.&  Arch.   &Params. & Time (s) & DAR (\%) &DAR (\%) &DAR (\%) &DAR (\%) &DAR (\%) &{DAR (\%)}\\    \midrule
our& ZINC    & SCAM     &HEM  &  5-3-44  &459k    & 2209.3   &\B91.50&\B	80.26&\B69.85&  66.32&\B92.25&\B80.04	(1.17)\\
   & CC      & SCAM     &HEM  &  5-3-44  &461k    & 1964.7   &  85.93&	77.86&	24.54&	41.85&	32.52&	52.54	(1.39)  \\
   & ZINC    & Max      &HEM  &  5-3-128 &457k    & 2013.1   &  90.57&	79.49&	7.74&	60.67&	78.01&	63.30	(0.80)  \\
   & ZINC    & SCAM     &CE   &  5-3-44  &459k    & 2228.3   &  70.54&	61.97&	42.16&	24.64&	31.54&	46.17	(2.05)  \\\midrule

std& CC      & Max      &CE   &  5-3-128 &458k    & 1890.5   &  89.49&	77.47&	5.01 &  47.25&	34.59&	50.76	(1.57)\\
   & ZINC    & Max      &CE   &  5-3-128 &457k    & 2021.1   &  66.04&	56.40&	46.21&	19.56&	29.47&	43.54	(1.35)\\
   & CC      & SCAM$\dagger$&CE &5-3-44  &461k    & 2017.0   &  84.86&	78.21&	12.88&	35.59&	25.58&	47.42	(1.32)\\
   & CC      & Max      &HEM  &  5-3-128 &458k    & 1957.8   &  88.40&	74.82&	11.73&\B68.89&	56.09&	59.99	(3.00)\\
\bottomrule\end{tabular}}\end{table*}

\subsection{Effect of Training Set-up} 
\label{sec-results_training_setup}

Networks built from the proposed building-blocks and size-matched standard CNNs were trained and tested on CIFAR10 using a number of combinations of learning rate, learning rate schedule, and optimiser (SGD \citep{Rosenblatt58}, RMSprop \citep{TielemanHinton12}, AdamW \citep{LoshchilovHutter19}, Lion \citep{Chen_etal23b}\footnote{\url{https://pypi.org/project/lion-pytorch/}}, Adan \citep{Xie_etal24}\footnote{\url{https://pypi.org/project/adan-pytorch/}}). Training was performed for 110 epochs with a batch size of 128. For the SGD optimiser, momentum was set to 0.9, all other hyper-parameters for all optimisers were left at their default values. For the Multi-Step scheduler the learning rate was decreased by a factor of 10 at epochs 100 and 105. Across these different training set-ups, the highest clean accuracy was produced by the proposed network architecture far more frequently than the standard architecture  (\cref{tab-training_setup}). This is surprising since these optimisers and their default parameter settings have been developed to work well with standard DNN architectures trained with CE loss.

The AdamW optimiser  combined with the One-Cycle scheduler was the most effective at training both architectures. This combination of optimiser and scheduler was the one used in all the other experiments reported in this paper. 
The result in \cref{tab-training_setup} confirm that the gap in performance between the proposed and standard architectures is unlikely to be due to the training set-up used. 

\subsection{Parameter Sensitivity Analysis}
\label{sec-results_sensitivity}

The proposed integration, pooling and loss functions introduce additional hyper-parameters. Results are not particularly sensitive to the values chosen for these hyper-parameters as shown in \cref{tab-results_DAR_sensitivity}, and as demonstrated by the fact that the same hyper-parameters have been used successfully with all the data-sets and network sizes used to produce the results in this article. Furthermore, it can be seen that for all the combinations of  hyper-parameter values tested the performance of the proposed network architecture exceeds that of the standard CNN across all metrics.

\subsection{Ablation Study}
\label{sec-results_ablation}
Good performance across all metrics requires all the proposed new components. Specifically, if the proposed network architecture is modified so that one component is replaced by the corresponding standard one, performance across all metrics drops, and for some metrics this  deterioration in performance is substantial (\cref{tab-results_DAR_ablation}, top panel). Similarly, if the standard network architecture is modified to replace one component by the corresponding proposed one, this is generally ineffective (\cref{tab-results_DAR_ablation}, bottom panel). The only exception is HEM loss, which when used to train networks with standard architecture increases the overall DAR by improving performance on adversarial and OOD rejection. This is consistent with the results in \citet{SpratlingSchuttHEM}. However, in all cases, including this one, the average DAR for a network using a single proposed component is always lower than that for a network containing all the proposed components.

\section{Conclusion}
\label{sec-conclusion}


We propose new architectural building blocks for DNNs to fix their persistent problem of producing confident, but wrong, predictions.
Firstly, given that NNs need to be able to recognise, and discriminate between, patterns, we design a new integration function that is better able to do this compared to the standard, linear, integration function.
Secondly, given that NNs need to be able to generalise to all class-preserving changes in appearance, not just spatial shifts, we design a pooling function that combines inputs across both space and a sub-set of channels.
Our empirical results show that equipping NNs with integration and pooling layers that are more capable of learning selective and tolerant representations, respectively, enables relatively shallow networks to perform competitively with much deeper existing NNs.
The proposed models are less prone to the generalisation and robustness issues that make the  behaviour of existing DNN diverge from that of humans.
We have also shown that for our new NNs, adversarial samples are no longer a distinct class of stimulus. Instead, like some other (more visible) image corruptions they result in low confidence predictions. This is a substantial, qualitative step forward in the handling of adversarial samples.



The lack of generalisation to adversarial samples suggest that the proposed models are too selective. This might be addressed in the future by developing improved methods of performing channel-wise pooling. Ideally, such a pooling function should be able to learn how to optimally assign channels to pools, or be able to add a channel to a pool dynamically during training. 
In this work we have not performed any direct comparisons of our proposed model to psycho-physical or neuro-biological data. Instead we have made claims about improved alignment to human visual perception based on reducing severe deficits in DNN behaviour that are apparent without such data: such as susceptibility to adversarial attacks and mistaking noise for an object. However, future work might usefully make such comparisons \citep{Geirhos_etal18,Schrimpf_etal20,Conwell_etal24}.
Further work might also more thoroughly evaluate the generalisation and robustness of STEAL Nets \citep{Geirhos_etal21}, assess their interpretability \citep{Doshi-VelezKim18}, and apply them to other tasks (with images and text). It would also be interesting to train the proposed models with more biologically-plausible methods such as those discussed in \cref{sec-bio_plausible_DNNs} and methods of self-supervised learning especially from video \citep{Maes_etal26,Aw_etal26,Daithankar_etal26} given this approach seems better aligned to human visual experience. Furthermore, the more ecologically valid training data augmentation methods and front-ends discussed in  \cref{sec-bio_plausible_DNNs} could also be combined with the proposed methods.
We also hope that our promising results will encourage more research on neural integration functions, which has been hugely under explored so far.


\section*{Acknowledgements}
The simulations were performed using the Luxembourg national supercomputer MeluXina and the HPC facilities of the University of Luxembourg \citep{Varrette_etal22}.
The authors gratefully acknowledge the LuxProvide and University of Luxembourg teams for their expert support.

\printbibliography[]

\clearpage
\appendix
\begin{center}
  \parbox{0.8\textwidth}{\centering \Large \bf Appendix\\\titletext}
  \end{center}

\section{Results in Terms of Standard Performance Metrics}

Here the same networks assessed in the main body of the paper are evaluated terms of  traditional metrics rather than DAR. The correspondence between tables in the main body and the appendix are as show in \cref{tab-DAR_conventional}.
\begin{table}[h]\centering
  \caption{Correspondence between tables in the main body and the appendix.}
  \label{tab-DAR_conventional}
  \begin{tabular}{ll}\toprule
    Evaluation using DAR & Evaluation using traditional metrics\\
    \midrule
    \cref{tab-results_DAR_all} & \cref{tab-results_conventional_all} \\
         \cref{tab-results_DAR_ImageNet_basic} &         \cref{tab-results_conventional_ImageNet_basic} \\
         \cref{tab-RobustBench_DAR}   &      \cref{tab-RobustBench_conventional}\\
         \cref{tab-results_DAR_sensitivity}  &       \cref{tab-results_conventional_sensitivity}\\
         \cref{tab-results_DAR_ablation}   &      \cref{tab-results_conventional_ablation}\\
    \bottomrule \end{tabular}\end{table}

\begin{table*}[tbp]\centering
  \caption{A comparison of the performance of STEAL networks composed of
    the proposed building-blocks (ZINC, SCAM, and HEM)
    with size-matched CNNs composed of standard building-blocks (CC, Max pooling, and CE).
    Bold text indicates the best performance on each metric for each training data-set. Coloured text indicates that the difference in performance is statistically significant.}
    \label{tab-results_conventional_all}
    \tabcolsep3pt
    \resizebox{\ifdim\width>\linewidth\linewidth\else\width\fi}{!}{%
      \begin{tabular}{llllllll *{4}{S[table-format=2.2(2)]}}\toprule
&Integrat. &Pooling &Loss  & Data-set & Network        &Num.    & Training & {Clean}  & {Corrupt} & {Advers.} & {OOD rejection} \\                     
&Func.      &Func.  &Func. &          & Architecture   &Params. & Time (s) & {Acc. (\%)} & {Acc. (\%)} & {Acc. (\%)} & {AUROC (\%)} \\ \midrule
our& ZINC   & SCAM   &HEM  & \multirow{2}{*}{MNIST}    & 2-5-20  & 44.0k  & 56.45   &\W99.29	(0.01)&\W   93.79	(0.15)&\W   1.98    (0.15)&\W 	99.17	(0.08)\\
std& CC     & Max    &CE   &                           & 2-5-160 & 44.2k  & 54.02   &  98.81	(0.07)& 	74.04	(1.66)& 	0.00    (0.00)&		96.22	(0.47)\\\midrule
our& ZINC   & SCAM   &HEM  & \multirow{2}{*}{FMNIST}   & 3-5-16  & 86.4k  & 160.02  &\W92.45	(0.16)&\W	62.42	(0.40)&	    0.00	(0.00)&\W	98.86	(0.24)\\
std& CC     & Max    &CE   &                           & 3-5-54  & 87.9k  & 157.75  &  90.92	(0.16)& 	45.85	(0.31)& 	0.00    (0.00)&		79.28	(2.21)\\\midrule
our& ZINC   & SCAM   &HEM  & \multirow{2}{*}{SVHN}     & 3-5-30  & 258k   & 233.51  &\W95.56	(0.25)&\W	69.93	(0.38)&\W	13.98	(0.25)&\W   97.83	(0.22)\\
std& CC     & Max    &CE   &                           & 3-5-96  & 262k   & 217.07  &  89.70	(0.23)& 	52.55	(1.40)& 	9.23	(0.27)&		85.16	(0.23)\\\midrule
our& ZINC   & SCAM   &HEM  & \multirow{2}{*}{CIFAR10}  & 5-3-44  &459k    & 2209.3  &\W92.32	(0.14)&\W	75.05	(0.36)& 	0.03	(0.01)&\W	95.21	(0.37)\\
std& CC     & Max    &CE   &                           & 5-3-128 &458k    & 1890.5  &  89.08	(0.28)& 	71.89	(0.05)&\F 	0.08	(0.02)&		83.67	(2.58)\\\midrule
our& ZINC   & SCAM   &HEM  & \multirow{2}{*}{CIFAR100} & 5-3-41  &667k    & 2072.6  &\W70.36	(0.19)&\W	45.04	(0.20)& 	0.03	(0.02)&\W	75.95	(0.63)\\
std& CC     & Max    &CE   &                           & 5-3-142 &677k    & 1913.1  &  63.66	(0.42)& 	40.71	(0.26)& 	0.03	(0.01)&		62.43	(1.79)\\\midrule
our& ZINC   & SCAM   &HEM  & \multirow{2}{*}{TinyIN}   & 7-3-60  &2.17M   & 39664   &\W49.97	(0.12)&\W	18.05	(0.25)&\W	0.97	(0.04)& 	54.35	(1.55)\\
std& CC     & Max    &CE   &                           & 7-3-200 &2.17M   & 24090   &  41.45	(0.31)& 	11.82	(0.18)&	    0.77	(0.07)&\B	55.31	(2.48)\\\midrule
our& ZINC   & SCAM   &HEM  & \multirow{2}{*}{ImageNet1k}&9-3-15-exp   &27.6M& 142908  &\B68.60&\B  		28.58&		    	0.00&\B		    	83.27\\
std& CC     & Max    &CE   &                            &9-3-425-exp  &27.2M& 196850  &  62.12&	 	    22.43&	 	        0.00&		    	80.33\\\midrule
\multicolumn{8}{r}{performance gap (our-std) averaged over data-sets} & +4.69   & +10.51 & +0.98  & +8.89  \\
\bottomrule \end{tabular}}\end{table*}

\begin{table*}[tbp]\centering
  \caption{Comparison of different DNN architectures trained on ImageNet1k for 100 epochs with basic training data augmentations. The first row shows result is for a network composed of the proposed components (ZINC, SCAM, and HEM). These results also appear in \cref{tab-results_conventional_all} and are repeated here for convenience. The remaining rows shows results for a variety of DNN architectures, ordered by increasing number of integration function layers.}
    \label{tab-results_conventional_ImageNet_basic}
    \tabcolsep4.5pt
     \resizebox{\ifdim\width>\linewidth\linewidth\else\width\fi}{!}{%
      \begin{tabular}{llllllll cccc}\toprule
 &Integrat.&   Pooling  &Loss       &Network                &Num.     &Time Per &Num. Int.   & Clean  & Corrupt & Advers. & OOD rejection \\
 &Func.    &   Func.    &Func.      &Architecture           &Params.  &Epoch (s)&Layers      & Acc. (\%) & Acc. (\%) & Acc. (\%) & AUROC (\%) \\ \midrule
our&ZINC   &   SCAM     &HEM        &9-3-15-exp      &27.6M    & 1429.1  &$9$   &  68.60&   28.58&	0.00&		83.27\\
   &CC/FC  &   Max      &CE         &AlexNet                &61.1M    & 1288.2  &$8$   &  55.49&	 22.89&	0.00&		74.61\\
   &CC/FC  &   Max      &CE         &VGG11                  &133M     & 1798.8  &$11$  &  69.72&   28.34& 0.00&       82.06\\
   &CC/FC  &   Max      &CE         &VGG16                  &138M     & 1604.0  &$16$  &  72.46&   32.03&	0.00&		82.85\\
   &CC/FC  &   Strided  &CE         &ResNet18               &11.7M    & 1268.2  &$21$  &  69.33&	 32.36&	0.00&		84.94\\
   &CC/FC  &   Strided  &CE         &ResNet50               &25.6M    & 1387.7  &$54$  &\B75.01& \B38.69&	0.00&		71.92\\
   &CC/FC  &   Strided  &CE         &ConvNeXt-tiny          &28.6M    & 1109.4  &$59$  &  73.38&	 36.09&	0.00&\B		85.36\\
   &FC/CC  &   None     &CE         &ViT-S16	            &22.1M    & 1101.4  &$74$  &  68.53&	 37.57&	0.00&		83.38\\
   &FC/CC  &   None     &CE         &ViT-B16	            &86.6M    & 2505.4  &$74$  &  65.56&	 35.22&	0.00&		81.28\\
   &CC/FC  &   Strided  &CE         &ConvNeXt-base          &88.6M    & 1778.9  &$113$ &  73.65&	 37.24&	0.00&		82.62\\
\bottomrule \end{tabular}}\end{table*}

\begin{table}[tp]
\begin{center}
  \caption{Comparison to models ranked highly on the RobustBench leaderboards. The column ``Origin'' indicates which leaderboard the model comes from. Note there are no RobustBench leaderboards for $l_2$ constrained adversarial attacks on CIFAR100 and ImageNet trained models. The best results for each column in each block are highlighted in bold.}
\label{tab-RobustBench_conventional}
     \setlength{\tabcolsep}{3.8pt}
     \resizebox{\ifdim\width>\linewidth\linewidth\else\width\fi}{!}{%
     \begin{tabular}{lcllC{21mm}C{21mm}C{21mm}C{21mm}}
        \toprule
      Origin     & Data-set &  Network      & Num.    & {Clean} &	{Corrupt}	& {Advers.} & {OOD rejection} \\
                 &            &  Architecture & Params. & {Acc. (\%)} & {Acc. (\%)} & {Acc. (\%)} & {AUROC (\%)} \\ \midrule

      our        &\multirow{14}{*}{\rotatebox{90}{CIFAR10}}     &5-3-225 &11.1M   & 94.60&	90.13&	0.15&\B			97.29\\[2mm]

RobustBench&   &WideResNet18-2 &\multirow{2}{*}{268M}        &\multirow{2}{*}{\B 96.56}&\multirow{2}{*}{\B93.49}&	\multirow{2}{*}{8.82}&  \multirow{2}{*}{94.25}\\
corruptions&   &\citet{Diffenderfer_etal21}&&&&& \\[1mm]
                                     
RobustBench&   &ResNet18 &\multirow{2}{*}{11.7M}               & \multirow{2}{*}{94.75}&	\multirow{2}{*}{90.45}&	\multirow{2}{*}{8.56}&	\multirow{2}{*}{88.92}
\\
corruptions&   &\citet{Kireev_etal22}&&&&&\\[1mm]
                                     
RobustBench&   &WideResNet70-16 &\multirow{2}{*}{267M}       & \multirow{2}{*}{95.54}&	\multirow{2}{*}{89.32}&	\multirow{2}{*}{68.06}&\multirow{2}{*}{91.34}\\
advers. ($l_2$)& &\citet{Wang_etal23}&&&&& \\[1mm]
                                     
RobustBench&   &WideResNet70-16 &\multirow{2}{*}{267M}       & \multirow{2}{*}{95.74}&	\multirow{2}{*}{88.97}&	\multirow{2}{*}{63.18}&\multirow{2}{*}{88.07}\\
advers. ($l_2$)&  &\citet{Rebuffi_etal21}&&&&&\\[1mm]
                                     
RobustBench&  &WideResNet94-16 &\multirow{2}{*}{366M}       &  \multirow{2}{*}{93.70}&	\multirow{2}{*}{85.61}&	\multirow{2}{*}{\B72.46}&\multirow{2}{*}{86.80}\\
advers. ($l_\infty$)& &\citet{Bartoldson_etal24}&&&&& \\[1mm]
                                     

RobustBench&  &WideResNet28-10 &\multirow{2}{*}{36.5M}       &  \multirow{2}{*}{92.16}&\multirow{2}{*}{84.43}&\multirow{2}{*}{68.67}&\multirow{2}{*}{83.93}\\
advers. ($l_\infty$)&  &\citet{Cui_etal23}   &&&&&\\[1mm]     

\midrule
      our        &\multirow{14}{*}{\rotatebox{90}{CIFAR100}}&5-3-225 & 11.1M  &77.86&	67.83&	0.17&\B			87.38\\[2mm]

RobustBench&  &WideResNet18-2 &\multirow{2}{*}{268M}       &   \multirow{2}{*}{\B79.92}&	\multirow{2}{*}{\B72.17}&	\multirow{2}{*}{7.74}&	\multirow{2}{*}{84.08}\\
corruptions & &\citet{Diffenderfer_etal21} &&&&&\\[1mm]

RobustBench&  &ResNet18 &\multirow{2}{*}{11.7M}               &  \multirow{2}{*}{77.59}&	\multirow{2}{*}{69.08}&	\multirow{2}{*}{0.61}&	\multirow{2}{*}{86.40}\\
corruptions & &\citet{Modas_etal22} &&&&&\\[1mm]     

RobustBench&  &ResNeXt29 32x4d &\multirow{2}{*}{6.9M}     & \multirow{2}{*}{78.90}&	\multirow{2}{*}{66.32}&	\multirow{2}{*}{0.09}&	\multirow{2}{*}{76.43}\\
corruptions&  &\citet{Hendrycks_etal20} &&&&&\\[1mm]   

RobustBench&  &WideResNet70-16 &\multirow{2}{*}{267M}      &  \multirow{2}{*}{75.23}&	\multirow{2}{*}{61.23}&	\multirow{2}{*}{\B43.77}&	\multirow{2}{*}{65.44}\\
advers. ($l_\infty$)&  &\citet{Wang_etal23} &&&&&\\[1mm]      

RobustBench&  &WideResNet28-10 &\multirow{2}{*}{36M}       &  \multirow{2}{*}{73.83}&	\multirow{2}{*}{60.20}&	\multirow{2}{*}{41.39}&	\multirow{2}{*}{58.85}\\
advers. ($l_\infty$)&  &\citet{Cui_etal23}   &&&&&\\[1mm]     

RobustBench&  &XCiT-L12 &\multirow{2}{*}{103M}             &  \multirow{2}{*}{71.20}&	\multirow{2}{*}{60.67}&	\multirow{2}{*}{40.73}&	\multirow{2}{*}{80.63}\\
advers. ($l_\infty$)&  &\citet{Debenedetti_etal23}  &&&&&\\

\midrule
      our        &\multirow{10}{*}{\rotatebox{90}{ImageNet1k}}& 7-3-40-exp & 26.3M & 69.04&	38.66&	0.00&		89.72\\[2mm]
RobustBench&  &DeiT Base     &\multirow{2}{*}{86.6M}         &  \multirow{2}{*}{\B82.40}&	\multirow{2}{*}{\B68.42}&	\multirow{2}{*}{5.65}&    \multirow{2}{*}{90.17} \\
corruptions&  &\citet{Tian_etal22} &&&&&\\[1mm]

RobustBench&  &ResNet50      &\multirow{2}{*}{25.6M}         &   \multirow{2}{*}{76.66}&	\multirow{2}{*}{52.71}&	\multirow{2}{*}{4.85}&	\multirow{2}{*}{82.11}\\
corruptions&  &\citet{Hendrycks_etal21} &&&&&\\[1mm]

RobustBench&  &Swin-L       &\multirow{2}{*}{197M}          &  \multirow{2}{*}{77.98}&	\multirow{2}{*}{55.05}&	\multirow{2}{*}{53.90}&	\multirow{2}{*}{89.92}\\
advers. ($l_\infty$) & &\citet{Liu_etal25}&&&&&\\[1mm]

RobustBench&  &ConvNeXt-L + ConvStem &\multirow{2}{*}{198M} &   \multirow{2}{*}{76.50}&	\multirow{2}{*}{53.16}&	\multirow{2}{*}{\B63.65}&\multirow{2}{*}{\B90.28}\\
advers. ($l_\infty$)&  &\citet{Singh_etal23} &&&&&\\[1mm]
\hline
      \end{tabular}}
\end{center}
\end{table}

\begin{table*}[tp]\centering
  \caption{A comparison of the performance of STEAL networks composed of the proposed building-blocks (ZINC, SCAM, and HEM) with size-matched CNNs composed of standard building-blocks (CC, Max pooling, and CE) when using adversarial training. Experiments performed using the CIFAR10 data-set. Bold text indicates the best performance on each metric for each training set-up.}
    \label{tab-results_conventional_CIFAR10_AT}
    \tabcolsep5pt
    \resizebox{\ifdim\width>\linewidth\linewidth\else\width\fi}{!}{%
      \begin{tabular}{lllllll cccc}\toprule
 &Integrat.&   Pooling  &Loss       &Network           &Num.     &AT    & Clean  & Corrupt & Advers. & OOD rejection \\             
 &Func.    &   Func.    &Func.      &Architecture      &Params.  &      & Acc. (\%) & Acc. (\%) & Acc. (\%) & AUROC (\%) \\ \midrule
our&ZINC   &   SCAM     &HEM        &5-3-225    &11.1M    &Weak  &\B94.60&\B	90.13&\B	0.15&\B	97.29\\
std&CC     &   Max      &CE         &5-3-640    &11.1M    &      &  92.05&	84.38&  	0.03&	96.04\\[1mm]
our&ZINC   &   SCAM     &HEM        &5-3-225    &11.1M    &Strong&\B80.70&\B	73.30&\B	51.66&	85.48\\
std&CC     &   Max      &CE         &5-3-640    &11.1M    &      &  75.03&	68.44&  	46.41&\B86.29\\
\bottomrule\end{tabular}}\end{table*}

\begin{table*}[tbp]\centering
  \caption{Evaluation of sensitivity to hyper-parameter values performed using the CIFAR10 data-set. The first row shows results obtained with the default hyper-parameter values and the last row shows results for a size-matched CNN built using standard components (these results also appear in \cref{tab-results_conventional_all} and are repeated here for convenience). The other rows show results when one hyper-parameter was changed from its default value while all other hyper-parameter remained unchanged. Bold text indicates the best performance on each metric.}
    \label{tab-results_conventional_sensitivity}
    \tabcolsep4pt
    \resizebox{\ifdim\width>\linewidth\linewidth\else\width\fi}{!}{%
      \begin{tabular}{lccccll *{4}{S[table-format=2.2(2)]}}\toprule
&\multicolumn{2}{c}{ZINC}&SCAM&HEM & Network        &Num.    &   {Clean}  & {Corrupt} & {Advers.} & {OOD rejection} \\                     
&$\lambda$&$\gamma$&$\kappa$&$\mu$ & Architecture   &Params. &   {Acc. (\%)} & {Acc. (\%)} & {Acc. (\%)} & {AUROC (\%)} \\ \midrule
our&2 & 0.1     &8  & 0.1         &  5-3-44  &459k      &  92.32	(0.14)& 	75.05	(0.36)& 	0.03	(0.01)& 	95.21	(0.37)\\[1mm]
   &1 &  0.1    &8  & 0.1         &  5-3-44  &459k      &\B92.54	(0.14)&	    73.47	(0.43)&	    0.00	(0.004)&	94.22	(0.88)\\
   &4 &  0.1    &8  & 0.1         &  5-3-44  &459k      &\B92.54	(0.12)&\B   76.10	(0.62)&\B   0.08	(0.01)&		95.06	(0.23)\\[1mm]  
   &2 &  0.05   &8  & 0.1         &  5-3-44  &459k      &  92.17	(0.13)& 	75.25	(0.31)& 	0.014	(0.002)&	94.91	(0.59)\\
   &2 &  0.2    &8  & 0.1         &  5-3-44  &459k      &  92.45	(0.12)& 	74.57	(0.62)&  	0.04	(0.02)&		92.78	(1.25)\\[1mm]
   &2 &  0.1    &4  & 0.1         &  5-3-63  &458k      &  92.35	(0.22)&	    74.83	(0.21)&	    0.03	(0.01)&\B	95.22	(0.23)\\
   &2 &  0.1    &16 & 0.1         &  5-3-30  &445k      &  92.18	(0.17)&	    74.58	(0.21)&	    0.03	(0.01)&		93.55	(0.81)\\[1mm]
   &2 &  0.1    &8  & 0.05        &  5-3-44  &459k      &  92.21	(0.10)& 	75.03	(0.25)& 	0.05	(0.01)&		94.52	(0.41)\\
   &2 &  0.1    &8  & 0.2         &  5-3-44  &459k      &  91.67	(0.21)& 	73.63	(0.32)& 	0.01	(0.01)&		94.51	(0.80)\\[1mm]
std&\multicolumn{2}{c}{CC}&Max&CE &  5-3-128 &458k      &  89.08	(0.28)& 	71.89	(0.05)&\B 	0.08	(0.02)&		83.67	(2.58)\\
\bottomrule\end{tabular}}\end{table*}

\begin{table*}[tbp]\centering
  \caption{Results for the ablation study  performed using the CIFAR10 data-set.  The top panel shows the effects of removing one of the proposed components (ZINC, SCAM, and HEM) and replacing it with the corresponding standard component (CC, Max pooling, and CE). The lower panel shows the effect of replacing components in a standard CNN with the proposed methods. $\dagger$\xspace In this condition, the parameter $\alpha$ used by SCAM was fixed to a value of one, as training failed when attempting to learn this parameter or when it was set to a higher value.}
    \label{tab-results_conventional_ablation}
    \tabcolsep3pt
    \resizebox{\ifdim\width>\linewidth\linewidth\else\width\fi}{!}{%
      \begin{tabular}{lllllll *{4}{S[table-format=2.2(2)]}}\toprule
&Integrat. &Pooling &Loss   &  Network         &Num.    & Training   & {Clean}  & {Corrupt} & {Advers.} & {OOD rejection} \\                     
&Func.       &Func.   &Func.  &  Architecture    &Params. & Time (s) & {Acc. (\%)} & {Acc. (\%)} & {Acc. (\%)} & {AUROC (\%)} \\ \midrule
our& ZINC    & SCAM   &HEM    &  5-3-44   &459k    & 2209.3   &\B92.32	(0.14)&\B	75.05	(0.36)& 	0.03	(0.01)&\B	95.21	(0.37)\\
   & CC      & SCAM   &HEM    &  5-3-44   &461k    & 1964.7   &  85.86	(0.16)&	    74.93	(0.40)&	    3.80	(0.38)&		75.80	(4.63)\\
   & ZINC    & Max    &HEM    &  5-3-128  &457k    & 2013.1   &  91.24	(0.24)&	    72.17	(0.35)&	    0.00	(0.002)&	92.64	(0.39)\\
   & ZINC    & SCAM   &CE     &  5-3-44   &459k    & 2228.3   &  68.33	(0.68)&	    52.63	(0.60)&	    0.01	(0.01)&		75.15	(2.59)\\\midrule

std& CC      & Max    &CE     &  5-3-128  &458k    & 1890.5   &  89.08	(0.28)& 	71.89	(0.05)& 	0.08	(0.02)&		83.67	(2.58)\\
   & ZINC    & Max    &CE     &  5-3-128  &457k    & 2021.1   &  62.76	(0.59)&	    46.85	(0.48)&	    0.00	(0.000)&	73.94	(1.02)\\
   & CC      & SCAM$\dagger$&CE &5-3-44   &461k    & 2017.0   &  82.92	(0.68)&	    73.58	(0.89)&\B   6.07	(1.67)&		74.47	(4.04)\\
   & CC      & Max    &HEM    &  5-3-128  &458k    & 1957.8   &  89.13	(0.25)&	    71.29	(0.50)&	    0.04	(0.01)&		89.00	(3.61)\\
\bottomrule\end{tabular}}\end{table*}

\end{document}